\documentclass{article}
\usepackage{arxiv}
\usepackage[T1]{fontenc}
\usepackage[utf8]{inputenc}\usepackage{amsmath}
\usepackage{amssymb}
\usepackage{amsfonts}
\usepackage{mathtools}

\usepackage{graphicx}
\usepackage{booktabs}
\usepackage{tabularx}
\usepackage{rotating}
\usepackage{float}
\usepackage{svg}
\usepackage{algorithm}
\usepackage{algpseudocode}
\usepackage{microtype}
\usepackage{nicefrac}
\usepackage{csquotes}
\usepackage[hidelinks]{hyperref}
\usepackage{cleveref}
\usepackage{doi}

\title{Context Steering: A New Paradigm for Compression-based Embeddings by Synthesizing Relevant Information Features}

\newif\ifuniqueAffiliation
\uniqueAffiliationtrue

\ifuniqueAffiliation 
\author{ \href{https://orcid.org/0000-0002-0268-849X}{\includegraphics[scale=0.06]{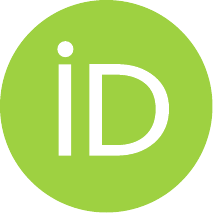}\hspace{1mm}Guillermo Sarasa Durán} \\
  Grupo de Neurocomputación Biol\'ogica\\
  Departamento de Ingenier\'ia Informática\\
  Escuela Polit\'ecnica Superior\\
	Universidad Aut\'onoma de Madrid\\
	Madrid, Spain \\
	\texttt{guillermo.sarasa@predoc.uam.es}\thanks{If the previous email doesn't
    work, use \texttt{guillermo.sarasaduran@gmail.com} instead} \\
	\And
	\href{https://orcid.org/0000-0003-0158-7969}{\includegraphics[scale=0.06]{orcid.pdf}\hspace{1mm}Ana Granados Fontecha} \\
  Escuela Polit\'ecnica Superior\\
  Universidad Carlos III de Madrid\\
	Madrid, Spain \\
  \texttt{agranado@inf.uc3m.es}\\
	\And
	\href{https://orcid.org/0000-0003-4053-099X}{\includegraphics[scale=0.06]{orcid.pdf}\hspace{1mm}Francisco de Borja Rodriguez Ort\'iz} \\
  Grupo de Neurocomputación Biol\'ogica\\
  Departamento de Ingenier\'ia Informática\\
  Escuela Polit\'ecnica Superior\\
	Universidad Aut\'onoma de Madrid\\
	Madrid, Spain \\
	\texttt{f.rodriguez@uam.es} \\
}
\else
\usepackage{authblk}
\usepackage{amsmath,scalerel}
\fi

\renewcommand{\shorttitle}{\textit{arXiv} Template}

\hypersetup{
pdftitle={Context Steering: A New Paradigm for Compression-based Embeddings by Synthesizing Relevant Information Features},
pdfsubject={cs.IT, math.IT}
pdfauthor={Guillermo Sarasa Durán},
pdfkeywords={Compression distance, NCD, NRC, Agglomerative clustering, Classification, Compressor, Kolmogorov Complexity},
}

\usepackage{cleveref}       
\usepackage{amsfonts}

\begin{document}
\maketitle
\begin{abstract}
  Compression-based dissimilarities (CD) offer a flexible and domain-agnostic means of
  measuring similarity by identifying implicit information through redundancies
  between data objects. However, as similarity features are derived from the
  data, rather than defined as an input, it often proves difficult to align with
  the task at hand, particularly in complex clustering or classification
  settings. To address this issue, we introduce ``context steering'', a novel
  methodology that actively guides the feature-shaping process. Instead of
  passively accepting the emergent data structure (typically a hierarchy derived
  from clustering CDs), our approach ``steers'' the process by systematically
  analyzing how each object influences the relational context within a
  clustering framework. This process generates a custom-tailored embedding that
  isolates and amplifies class-distinctive information. We validate this
  supervised context-steering strategy using Normalized Compression Distance
  (NCD) and Relative Compression Distance (NRC) combined with hierarchical
  clustering, and evaluate the learned embeddings through both classification
  performance and cluster-quality metrics. Experiments on heterogeneous
  datasets—from text to real-world audio— show that the proposed approach yields
  robust task-oriented embeddings from compression dissimilarities, moving from
  traditional transductive uses of distance matrices to an inductive
  representation that can be applied to unseen data.
\end{abstract}

\hypersetup{
  pdftitle={Context Steering: A New Paradigm for Compression-based Embeddings by Synthesizing Relevant Information Features},
  pdfsubject={q-bio.NC, q-bio.QM},
  pdfauthor={Guillermo Sarasa Durán; Ana Granados Fontecha; Francisco de Borja Rodriguez Ort\'iz},
  pdfkeywords={Compression distance, NCD, NRC, Agglomerative clustering, Classification, Compressor, Kolmogorov Complexity}
}

\section{Introduction}

Compression-based dissimilarities have long been valued for their powerful
domain-agnostic capabilities, deriving similarity measures directly from data
without the need for manual feature engineering
\cite{cilibrasiClusteringCompression2005}. However, despite their theoretical
appeal, their practical application often faces significant challenges
\cite{cebrianCommonPitfallsUsing2005}. On the one hand, highly specialized
transformations can provide strong clustering results but suffer from limited
generality \cite{oshaughnessyMalwareFamilyClassification2021,hurwitzNeuralNormalizedCompression2024}
and poor adaptability to new problems. On the other hand, generic
transformations--such as directly clustering raw distance matrices--often
yield inconsistent performance, heavily depending on the intrinsic properties of
the dataset at hand \cite{borbelyNormalizedCompressionDistance2015}. This
limitation has historically restricted the broader adoption of compression-based
methods in complex or heterogeneous domains.

In this work, we propose a supervised context-steering approach that uses
compression-based clustering to build a task-oriented embedding aligned with the
problem objective, enhancing the utility of compression dissimilarities for both
clustering and downstream analysis. Our methodology is designed to overcome the
traditional dichotomy between small, highly controlled experimental settings and
broad, unsupervised approaches that depend solely on the intrinsic quality of
the compression dissimilarities. Instead of relying on the inherent separability
present in the dataset, we introduce a mechanism that actively “steers” the
notion of separability towards structures of interest (\textit{e.g.}, classification
problems, anomaly detection) by identifying and retaining those elements whose
induced relational context best matches the intended objective.

The core of our method lies in taking advantage of the context that emerges as
an outcome of applying clustering techniques over compression dissimilarities. As the
combination of compression dissimilarities and clustering produces a specific
``context'' (as the meaning extrapolated from a structure such as a hierarchical
tree) among the objects, adding or subtracting objects to this set also alters
the overall structure. Building on this concept, we analyze how each object
contributes to this context by isolating and testing the role that it plays.
This process creates a methodology to generate custom-tailored contexts from
data objects alone.

The remainder of the paper is organized as follows. Section~\ref{sec:background}
reviews the key concepts behind compression dissimilarities and outlines previous
attempts to bridge the gap toward clustering. Section~\ref{sec:methodology}
introduces our approach to context orientation, including a detailed analysis of
the mechanisms in hierarchical clustering that we exploit.
Section~\ref{sec:materials_methods} describes the datasets and the specific
compression models used. Section~\ref{sec:experiments} presents a thorough
experimental evaluation intensively testing the capabilities of our method
across multiple scenarios with different levels of complexity.
Section~\ref{sec:discussion} discusses the experimental findings, with emphasis
on technical insights and interpretation. Section~\ref{sec:conclusions}
summarizes the main contributions of the work and outlines its broader
implications. Additionally, for simplicity, Section~\ref{sec:notation} compiles
the notation and terminology used throughout the paper.

\section{Background and Related Work}
\label{sec:background}

In this section, we first review compression-based dissimilarities, which form the
theoretical foundation of our approach. We then discuss how methods like
clustering have been used to adapt these distances for various tasks,
highlighting the issues and advantages of applying these methods. Building on
this background, we introduce our perspective on how these adaptation techniques
are applied, and conclude with a brief description on how these methods form the
basis of the methodology developed and evaluated throughout this work.

\subsection{Compression Dissimilarities}
\label{subsec:compression}

In the field of data analysis and information theory, compression-based dissimilarities offer a powerful framework for quantifying similarity between objects.
By leveraging the compressibility of data as a basis for capturing information,
these methods provide a domain-agnostic way to compare objects, measuring shared
redundancies and structural relationships. Unlike conventional distance
measures, which often require predefined feature spaces, compression dissimilarities
derive their comparisons directly from the data, making them versatile tools for
diverse applications. The two formulations used in this work are the well known
Normalized Compression Distance (NCD) \cite{cilibrasiClusteringCompression2005} and the Normalized Relative Compression
 (NRC) \cite{Pratas2016CompressionGenomicData,pratasComparisonCompressionBasedMeasures2018} with distinct properties and applications, as explored
below.

\paragraph{The Normalized Compression Distance} (NCD) stands as one of the most
recognized and widely utilized formulations for measuring similarity through
data compression. The basic notion behind it is that, if two compressed files
occupy more space when stored separately than together (\textit{i.e.} compressed
while concatenated) they should share some information.

The NCD \cite{cilibrasiClusteringCompression2005,liSimilarityMetric2004a}is
defined over a standard compression algorithm $C(x)$, which computes the
compressed length of a given string $x$. Its most common formulation is
expressed as:

\begin{equation}
\text{NCD}(x,y) = \frac{C(xy) - \min\{C(x), C(y)\}}{\max\{C(x), C(y)\}}, \label{eq:ncd}
\end{equation}

\noindent where $C(xy)$ denotes the compressed size of the concatenation of $x$ and $y$.
This formulation quantifies the shared information between $x$ and $y$ relative
to their individual complexities, making it a versatile and effective tool in
numerous applications across data analysis and information theory.

The NCD has been applied to numerous research areas due to its parameter-free
nature, broad applicability, and robust performance. For example, it has been
successfully employed in diverse contexts such as text and document
analysis~\cite{WOS:000303281800004,granadosReducingLossInformation2011,jiangLessMoreParameterFree2022,hurwitzNeuralNormalizedCompression2024},
time-series and biomedical signal
interpretation~\cite{pascarellaFunctionalBalanceRest2024,sarasaAlgorithmicClusteringBased2019},
genomic sequence
classification~\cite{aliUniversalNonparametricApproach2024,WOS:000786894000001},
music and audio-based
categorization~\cite{gonzalez-pardoInfluenceMusicRepresentation2010,sarasaAutomaticTreatmentBird2018},
image comparison~\cite{WOS:000426793900022}, video activity
recognition~\cite{sarasaCompressionBasedClusteringVideo2018}, and even in
structural domains like graph
matching~\cite{gilliozNormalizedGraphCompression2025}. Its high resistance to
noise and independence from domain-specific assumptions have made it a
compelling choice across a variety of data types and tasks. However, despite its
strengths, the practical strategies used to leverage the NCD in clustering or
classification settings have remained relatively limited. The most common
approaches in the literature\footnote{There are many works in the literature
  that provide \textit{ad-hoc} methods to use compression dissimilarities, hence
  here we focus in those that are most widely applied among works as reference
  to a general procedure.} is to rely on either the quartet method or k-nearest
neighbors, while more tailored adaptations tend to lack generalizability,
hindering the development of reusable and context-aware methodologies.

\paragraph{The Normalized Relative Compression} measure (NRC), in contrast to
the previous measure, is rarely employed as a measure of similarity.

The basic notion of this measure relies on compressing two files without needing
to store any information from one of them. Thus, compressing one relative to the
other. The NRC relies on a specialized class of compressors $C(x\Vert y)$
capable of performing exclusive compression of $x$ with respect to $y$. Its
standard formulation
\cite{WOS:000475304200073,WOS:000507375900064,WOS:000436275400002,castroFHRBiometricIdentification2018,carvalhoExtendeDalphabFinitecontext2018}
is given by:

\begin{equation}
\text{NRC}(x\Vert y) = \frac{C(x\Vert y)}{|x| \log(|A|)}, \label{eq:nrc}
\end{equation}

\noindent where $|A|$ represents the size of the alphabet\footnote{In some works in the
  literature $|A|$ is ignored, as a result of measuring distances
  between objects with a common alphabet.} of $x$. While its theoretical
properties diverge from those of the NCD, the motivation for employing the NRC
in this study arises from its alignment with our goal of using distances as
references (detailed in Section~\ref{subsec:clustering}). Given this context,
including the NRC ensures a more rigorous approach, as it provides a logical and
consistent framework for treating compression dissimilarities in this manner.

Despite its more limited presence in the literature, the NRC has been
successfully applied across several research domains, including biomedical
signal processing
\cite{ramosImpactDataAcquisition2021,carvalhoExtendeDalphabFinitecontext2018}
and biological sequence analysis
\cite{carvalhoImpactAcquisitionTime2017,WOS:000436275400002}, authorship
attribution and textual data exploration
\cite{pinhoAuthorshipAttributionUsing2016,WOS:000475304200073}, as well as
comparative evaluations in structured or computational reconstruction settings.
These varied applications demonstrate its practical utility as a similarity
distance. While the NRC does not achieve an absolute measure of similarity, it
has proven to be a robust and effective alternative in scenarios where relative information
plays a meaningful role.

\subsection{From Compression Dissimilarities to Data Clusters}
\label{subsec:clustering}

While the Normalized Compression Distance (NCD) provides a powerful and
universal measure of pairwise similarity, its most common application involves
assembling a square matrix of distances over all the initial objects. This
yields a dense $N^2$ matrix of values that often cannot be directly interpreted.
Even considering the symmetry of the NCD matrix, which reduces the number of
unique entries to $\frac{N(N+1)}{2}$, this reduction does not sufficiently
alleviate the computational burden to make it practical in most scenarios.
Furthermore, compression-based similarities are not guaranteed to reflect shared
dominant features across all object pairs; instead, each distance may reflect a
different particular behavior between each pair of objects, complicating global
interpretation. The basic notion of this concept is that, as the capability of a
measure of finding redundancies grows, its specificity decreases, creating
scenarios where every pair of objects is ``similar'', but each one for different
reasons. This phenomenon is explicitly discussed in
\cite{cilibrasiClusteringCompression2005}, where the authors state that the NCD
might reflect the most dominant feature shared by each pair of objects
individually, which may vary significantly across the dataset.

To address this, the original authors of the NCD dismiss simple approaches such
as Minimum Spanning Trees (MST), which fail to capture the nuanced dependencies
in the distance matrix. Instead, they introduce a principled alternative based
on \emph{quartet topologies}, which consider all subsets of four elements and
define trees as consistent when they embed as many of these local structures as
possible.

This formulation leads to the \emph{Minimum Quartet Tree Cost} (MQTC) problem,
which seeks to find the tree that best represents the dissimilarity matrix by
minimizing a global cost function over all consistent quartets. For a given tree
\( T \), the total cost is defined as the sum of distances associated with the
quartet topologies embedded in the tree. This is then normalized to produce a
standardized benefit score \( S(T) \in [0,1] \), which reflects the degree to
which \( T \) captures the minimal-cost quartet topologies across all possible
combinations.

This approach represents a foundational method for clustering NCDs into a tree
structure. Despite its theoretical soundness, solving the MQTC problem in
practice presents serious obstacles. Chief among these is the
\(\mathcal{O}(N^3)\) computational complexity, where \(N\) is the number of
elements, as well as the randomized initialization that offers no guarantee of
convergence to an optimal solution \footnote{The original authors suggest a
  practical upper limit of approximately 40 objects, beyond which tree quality
  tends to degrade.}. In previous works
\cite{sarasaCompressionBasedClusteringVideo2018,sarasaAutomaticTreatmentBird2018,sarasaAlgorithmicClusteringBased2019,WOS:000425242400021}
we observed that this lack of determinism can significantly prolong run times
and sometimes necessitate recomputations. Consequently, we advocate for standard
Agglomerative Hierarchical Clustering
\cite{wardHierarchicalGroupingOptimize1963,HierarchicalClustering2011} (AHC) as
a more efficient and stable approach, benefiting from \(\mathcal{O}(N^2)\)
complexity through optimized linkage algorithms (pseudocode included in
Algorithm \ref{alg:hca}).

\subsubsection{From NCD to NRC: Overcoming Challenges in Hierarchical Clustering}

Notwithstanding its benefits, AHC poses several challenges when directly paired
with compression-based dissimilarities such as the NCD. Its core mechanism
typically relies on an initial pairwise distance matrix, and certain linkage
strategies (e.g., Ward’s) explicitly require Euclidean distances. While some
linkage criteria allow for the use of precomputed distances, using NCDs in this
manner often produces trees that fail to capture meaningful structures within
the data. This limitation arises from AHC’s iterative ``chain'' approach:
whenever two clusters merge, distances to remaining clusters must be
recalculated in order to proceed. Hence, AHC uses aggregation functions (such as
average, centroid or ward) to measure these new distances. Given that NCDs tend
to express a wide range of similarities among different files (as they uncover
relations beyond the intended purpose of the analysis), the aggregation approach
tends to underperform with them when used as precomputed distances.

\begin{figure}[ht!]
  \centering
  \includesvg[width=\textwidth]{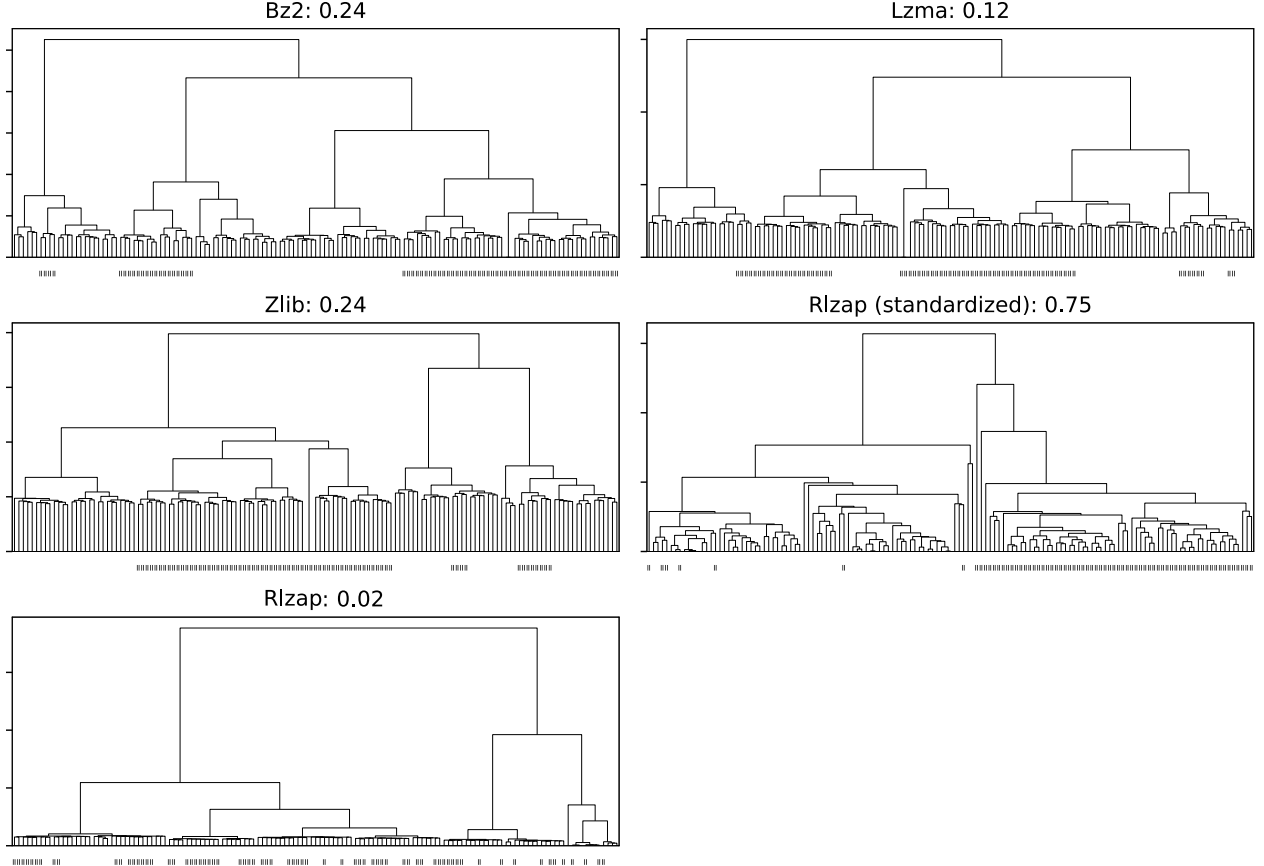}
  \caption{\textbf{Hierarchical dendrograms obtained from the \emph{Easy} dataset
    experiment}, each containing 138 samples from two classes (one indicated by a
    small double vertical line (``$\Vert$''), and the other by an empty space
    (`` '')), for different compressors. The title of each plot reports the
    silhouette coefficient along with the compression method used (e.g., bz2,
    lzma, zlib, rlzap: non-standardized and standardized. Every compressor is
    using the NCD to compute distances, with the exception of both rlzap cases,
    that use the NRC formulation. All dendrograms were generated using Ward’s
    linkage and Euclidean distances. The silhouette coefficients are computed
    using the height of the tree as distance between objects. For simplicity,
    heights have been removed from the figures.}
  \label{fig:hctrees_poecraft}
\end{figure}

A viable workaround involves treating NCD distances as feature vectors rather
than direct clustering inputs. Specifically, using each object’s row of NCD
values into a ``signature vector'' provides a more context-aware (as in including
the overall structure of dissimilarities of the set into the distance between
two objects) representation for hierarchical clustering. However, this
feature-based approach introduces additional complexities. NCD does not
uniformly reflect common references: two NCD computations sharing a reference
object (\textit{e.g.}, \(\text{NCD}(x,z)\) and \(\text{NCD}(y,z)\)) may capture
differing underlying redundancies, thereby complicating the interpretability and
coherence of resulting features (\textit{i.e.}, the values of each column).

To overcome this limitation, we add to our experiments the \emph{Normalized
  Relative Compression} (NRC) as a more principled alternative. Since NCDs are
distances defined between a pair of objects, it could be problematic to try to
fit together each set of distances that share a common object (\textit{i.e.}
each $\text{NCD}(X, y)$, where $X$ represents a set of objects, and $y$ a single
object. With NRCs, in contrast, this step is automatic as it is inherently
asymmetric and explicitly quantifies the \emph{exclusive redundancy} of a target
object with respect to a reference. For example, while \(\text{NCD}(x,y)\)
measures shared similarity, \(\text{NRC}(x\Vert y)\) captures how much of \(x\)'s
content is explainable solely by \(y\). 

Nevertheless, NRC values can still be sensitive to disparities in object sizes,
as the target’s original size directly influences its resulting measure.

And while the NCD approaches said differences by normalizing with respect to the
larger compressed object's size, the NRC only considers the original size of the
base object (\textit{i.e.}, the object from which the distance is measured).
This minimal size compensation, opens the door to possible issues that may
distort distances between objects with remarkable size differences. We therefore
incorporate a row-wise standardization procedure to normalize these values
exclusively for the NRC case. As we assume that the size differences between
objects and references, affect each object's comparisons (\textit{i.e.}, each
observation feature vector) in the same way across all NRCs. When possible, the
statistics (mean and standard deviation) for the standardization, are derived
from an external reference dataset (related to the problem) rather than from the
experimental dataset itself. As we wanted to ensure that the normalization was
focused on removing the size characteristics produced by the original object's
size, not the current references' sizes used for that analysis.

Our experimental findings (see Figure \ref{fig:hctrees_poecraft}) confirm that
this row-wise standardization approach, significantly improves class separability
when performing AHC on the standardized NRCs compared to the original ones,
thereby paving the way for improving future analysis over NRCs. The effect of
this normalization is shown in depth in the experiments of
Section~\ref{sec:experiments} and discussed further in
Section~\ref{sec:discussion}.

\section{Context Steering: Reshaping Compression Dissimilarities into Context-Based Features}
\label{sec:methodology}

Motivated by the limitations discussed above, we develop an approach—Context
Steering—that starts from a precomputed compression-dissimilarity matrix $\mathcal{M}$, whose
entries quantify how dissimilar pairs of objects are (for instance, using NCD or
NRC). From this matrix, our aim is to build a context-based embedding $V$, where
each object is represented by a feature vector that can be used by standard
clustering and classification methods. Rather than passively accepting the
structure induced by $\mathcal{M}$, our approach actively reshapes the context among
objects by selecting and weighting specific samples and features, favouring
those that best reflect the task of interest. In the rest of this section, we
formalize this idea through the notions of behavior similarity and structural
alignment, and describe how they are used to construct the final embedding.

As mentioned before, the Normalized Compression Distance (NCD) has long been valued for its
domain-agnostic nature, deriving similarity measures solely from the data itself
without relying on knowledge of the specific application domain. While this
generality affords unbiased similarity assessments between pairs, it also means
that direct one-to-one comparisons can be difficult, as the dominant features
considered for each distance may vary between pairs of objects.
Traditionally, both embedding techniques or clustering algorithms have been used
to ``balance'' similarities between sets of objects, trying to find a common
structure that could fit as much similarities as possible. The organization
shown in these structures, is referred in this work as ``context'' of
a set of data objects. For example, the quartet method described above, attempts
to represent NCDs in a tree structure, minimizing an error function. However,
such transformations can be highly dependent on the properties of the dataset
(\textit{e.g.} number of objects), are transductive (\textit{i.e.}, not
generalizable to unseen examples), and may struggle with class imbalance or
highly heterogeneous data (\textit{e.g.} trying to fit many different behaviors
under the same tree) — particularly when the data representation does not align
well with the chosen compressor.

Our experiments indicate that while applying transformations to compression dissimilarities can uncover meaningful structures in simpler cases, their efficacy
diminishes as complexity increases. In this sense, two key limitations of compression dissimilarities
become evident: a growing dataset size, which makes pairwise computations
structurally challenging (on top of being increasingly expensive); and a larger
number of classes, which reduces separability and can blur distinctions in both
distance-based clustering and feature-space embeddings. Consequently, these
transformations may fail to preserve meaningful clusters, highlighting the need
for additional help to transform distances into useful structures.

\begin{figure}[ht]
  \centering
  \includesvg[width=\textwidth, inkscapearea=drawing, inkscapelatex=false]{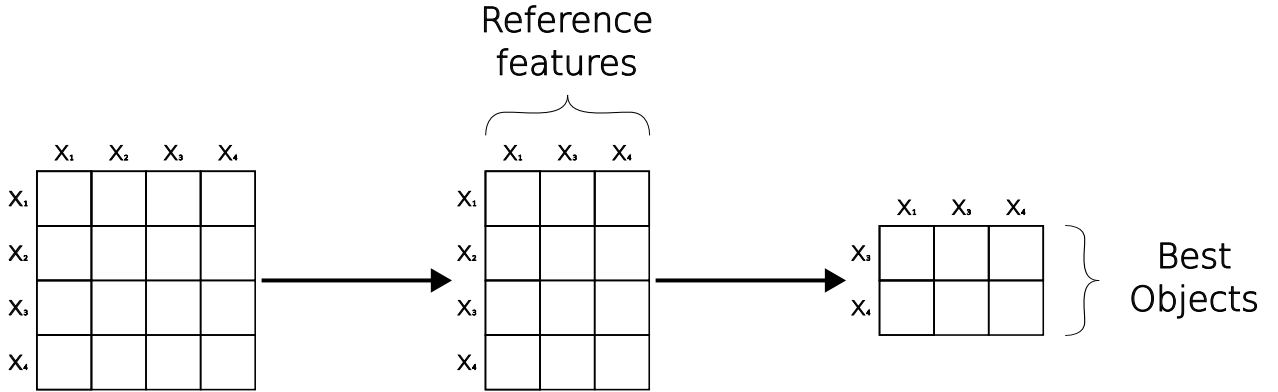}
  \caption{\textbf{Illustration of the two different methods of object selection} applied
    in our method. Initially, the matrix is symmetric and square, where both
    rows and columns represent the same set of objects ($x_1$, $x_2$, $x_3$, and
    $x_4$), and each cell encodes the pairwise compression dissimilarity between
    them. Applying AHC over this matrix, will sort the objects into a single
    hierarchical tree, hence producing what we understand as the context of the
    set. In our method, this matrix is reinterpreted so that each row becomes
    a sample, and each column a feature, with feature values corresponding to
    the distance to each particular reference object. In the center of the
    figure, a subset of columns is selected (e.g., $x_1$, $x_3$, $x_4$),
    following a certain criterion analogous to feature selection in machine
    learning. In the right part of the figure, a sample selection is applied
    row-wise, keeping only those samples (e.g., $x_3$ and $x_4$) that are
    well-aligned with a second criterion. This two-step process refines both the
    representational space (the features that define the profile
    or behavior of each sample) and the sample set (the set of
    samples to fit together into a single tree). Allowing for a different
    representation to emerge from the application of AHC over the data. And,
    thereby, defining a new context to work with in future steps.}
  \label{fig:contextual_transformation}
\end{figure}

To address these limitations, we propose a method that uses the emergent context
given by the combination of clustering compression dissimilarities, as a validation
mechanism to ``\textit{steer}'' the analysis towards a given goal.

Rather than relying on a global representation of the dataset (such as a single
tree), our method constructs local representations and then manipulates them to
emphasize those distinctions relevant to the target. In this setting, each
object is described by its distances to other objects in the set. We refer to
this array of values as a ``\textit{behavior}'' profile—a vector of its
compression dissimilarities to a set of reference elements. In a sense, defining how
each object behaves. These references will act as features in the clustering
process, which makes the choice of reference set analogous to feature selection
in a conventional machine learning problem: by choosing which columns to include
(\textit{i.e.}, which reference objects to consider), we influence which patterns
are deemed relevant by the clustering algorithm, without removing the
corresponding rows from the dissimilarity matrix (as we detach the relation between
rows and columns that distance matrices have). As a visual aid, a basic notion
of this concept is represented in the center of
Figure~\ref{fig:contextual_transformation}.

This form of selective comparison allows us to construct contextual spaces
specifically tailored to the problem's structure. For instance, in a binary
classification setting, an ideal reference set would help distinguish the
internal cohesion of each class and emphasize their boundaries. An obvious
following step would be to, once the references are selected, observe how each
object behaves in this new space. As a result, two types of samples will be
obtained for each class: those that remain grouped together within their class
and those that do not. And as this new space is defined over measures of
information similarity, we could define the former as the samples that contain
``relevant'' information (as the information capable to distinct one class from
the other) for the current context, while the latter contain, in the best case,
the ``shared'' information. As it could also contain information not
identifiable among its class. Looking at Figure~\ref{fig:hctrees_poecraft}, this
distinction is exemplified by the second row, second column tree, where almost
every object is well-contained within its class. In contrast, the third row,
first column tree better represents the case where no sample fits into its
class.

Crucially, this also provides a second layer of selection, this time over the
observations themselves. Once a context has been established via column
selection, we can assess which samples fit cohesively within that context.
Retaining only those well-aligned observations does not just refine the dataset,
it also enhances the structural consistency of the remaining samples. This
reinforces task-relevant relationships and suppresses noise or ambiguity in the
induced clustering space. Which can prove to be useful in the creation of
embeddings to fit new data into a custom tailored space for the previously
defined context. This will be explained in depth in
Section~\ref{subsubsec:first_step}. As a visual aid, this concept is represented
on the right side of Figure~\ref{fig:contextual_transformation}.

\subsection{From Hierarchical Clustering to Context Selection}
\label{subsec:hca_mechanisms}

Agglomerative Hierarchical Clustering
\cite{wardHierarchicalGroupingOptimize1963,HierarchicalClustering2011}
operates through a clear sequence of steps. Initially, the method requires
either a metric to compute distances between samples, and/or a precomputed
matrix of distances. Based on these distances, AHC iteratively merges the two
most similar samples or clusters according to a predefined linkage criterion
(such as single, complete, or average linkage). After each merge, the distance
matrix is updated to reflect the newly formed cluster, and the process repeats
until all objects are combined into a single cluster, forming a comprehensive
hierarchy. A detailed description of the complete procedure can be found in
Algorithm~\ref{alg:hca}.

\begin{figure}[ht!]
  \centering
  \includesvg[width=0.75\textwidth, inkscapearea=drawing, inkscapelatex=false]{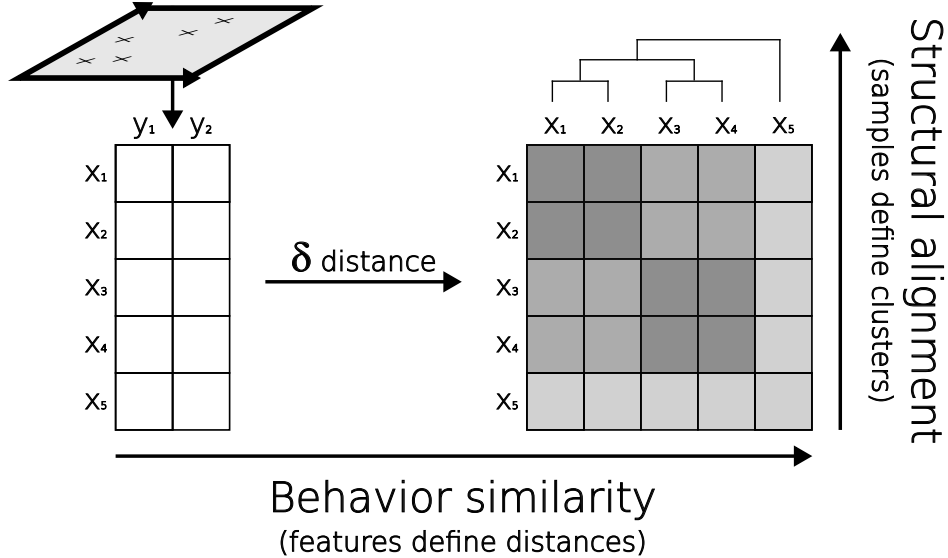}
  \caption{ \textbf{Basic illustration of the two key phases in Agglomerative
      Hierarchical Clustering}. For this example, a $R^2$ space is defined by
    the features $y_1$ and $y_2$. The different points are defined by $x_i$. In
    the left part of the figure, we have the matrix of observations that defines
    the position of every $x_i$ point in $R^2$. Then, to compute the matrix of
    distances that AHC will use, a certain distance function $\delta$ is used.
    The second step is to iteratively aggregate samples into clusters
    (recomputing distances for each new cluster) until only one cluster remains.
    If one were to change the features used in the first step, for example by
    introducing a $y_3$, the distance between samples will change. This is what
    we refer as ``behavior similarity'' as the features used in the analysis
    bias the distances (or dissimilarities) used in the clustering. The same can
    be done for the samples used, for example by removing $x_2$ and $x_4$,
    affecting how the final tree is assembled. In this work, we refer to this
    factor as ``structural alignment'' as the samples used in the analysis
    define the final structure of the tree.}
  \label{fig:hca_steps}
\end{figure}

Within this general framework, two specific stages of AHC offer natural points
of intervention for introducing context into the process. As a visual aid,
Figure~\ref{fig:hca_steps} illustrates these two stages using a simple example.

The first element concerns the computation of distances. In our framework, each
object is represented by a ``\emph{behavior profile}'': the vector of its compression dissimilarities to a set of reference objects. By selecting which references (columns)
to keep, we change the space in which similarities are evaluated: removing
references that create disagreement between two objects can make them appear
closer, while preserving discriminative references sharpens their separation. We
refer to this capacity to shape similarity relationships through feature
(column) selection as \emph{behavior similarity}; in practice, we compare behavior
profiles using Euclidean distance.

The second element concerns the organisation of samples into clusters. Even with
a fixed linkage criterion, the composition of the sample set strongly influences
the resulting tree: removing samples that create bridges between classes or
disrupt intended groupings simplifies the structure and stabilises the
partitions of interest. We refer to this control over which samples (rows)
participate in the clustering as \emph{``structural alignment''}: by including samples that
support the target partition and excluding those that contradict it, we steer
the overall tree structure toward the task at hand.

These two intervention points--\emph{behavior similarity} through feature selection
and \emph{structural alignment} through sample selection--provide the base (the first
step) for the method we develop next. By exploiting them systematically, we
transform compression-based dissimilarities into context-driven embeddings that enhance
clustering performance without altering the underlying data representations.

\subsection{Introducing Context in Clustering by Compression}
\label{subsec:approach}

Our proposed methodology is organized into three main steps, with an optional
refinement of the Euclidean layer used in the first step. Conceptually, the
method constructs class-wise hierarchical structures that based on
full multi-class context, without forcing all classes
into a single global dendrogram. This is achieved by (i) comparing samples
through their behavior profiles (rows of the compression-dissimilarity matrix)
and (ii) restricting the linkage phase of hierarchical clustering to within-class
samples only. Once class-wise dendrograms are obtained, we extract the most
coherent intra-class groups and synthesize an inductive embedding from
those groups by aggregating redundant compression-dissimilarity features.

The steps are summarized below (see Figure~\ref{fig:method_overview}):
\begin{itemize}
  \item \textbf{Step 1: Full-context behavior geometry and class-wise linkage.}
        Build a behavior similarity comparing full behavior profiles by means of
        the Euclidean distance (Step~1.a), and then run hierarchical linkage
        separately within each class using those full-context features
        (Step~1.b), producing one dendrogram $\mathcal{T}_k$ per class.

  \item \textbf{Step 2: Identify coherent groups.} Convert each $\mathcal{T}_k$
        into a flat partition by maximizing the cluster separability (silhouette
        coefficient), rank the resulting clusters by coherence, and retain the
        most coherent ones via a selection operator $h(\cdot,\tau)$. The
        retained clusters are collected into a global set of coherence groups
        $G=\{G_j\}$.

  \item \textbf{Step 3: Aggregated-cluster embedding and context feature
        filtering.} For each coherence group $G_j$, select a small set of inner
        references $R_j$ (Step~3.a), compute a reference-conditioned weighted
        representation of the group (Step~3.b), and aggregate it into scalar
        coordinates per sample (Step~3.c). Concatenating these coordinates
        yields the final embedding $\phi_f(\cdot)$, which can be applied to
        unseen samples using train-only objects.

\end{itemize}

The notation used throughout this section is summarized in
Section~\ref{sec:notation}. We now formalize each step.

\subsubsection{Step 1: Introducing Context into Intra-Class Distances (with optional distance refinement)}
\label{subsubsec:first_step}

Let $X=\{x_i\}_{i=1}^{N}$ be a set of objects with labels
$Y=\{y_i\}_{i=1}^{N}$, where $y_i\in\{1,\dots,K\}$. Let
$\mathcal{M}\in\mathbb{R}^{N\times N}$ be the full pairwise compression-dissimilarity matrix,
with entries $m_{ij}=d(x_i,x_j)$, where $d(\cdot,\cdot)$ denotes a compression dissimilarity operator (instantiated in our experiments as NCD or NRC). We interpret
each row $\mathcal{M}[p,:]$ as the behavior profile of object $x_p$, i.e., a vector
of distances-to-references against $X$.

\paragraph{Step 1.a: Behavior similarity geometry (Euclidean distance).}
We define a behavior similarity distance between objects $p$ and $q$ as the
Euclidean distance between their behavior profiles:
\begin{equation}
  E_{p,q}=\delta_{\text{eucl}}(p,q) = \Big\lVert \mathcal{M}[p,:]-\mathcal{M}[q,:]\Big\rVert_2 = \sqrt{\sum_{i=1}^{N} (m_{p,i}-m_{q,i})^2},
  \label{eq:euclidean}
\end{equation}

\noindent yielding an Euclidean matrix $E\in\mathbb{R}^{N\times N}$, where $E_{p,q}=E(p,q)$.
Optionally, we may replace $\delta_{\text{eucl}}$ with a blockwise alternative
$\delta_{\text{alt\_eucl}}$ (Section~\ref{subsubsec:add_step}) to mitigate
imbalances across class-specific feature blocks in $\mathcal{M}$.

\paragraph{Step 1.b: Class-wise linkage under full-context features.}
For each class $k$, let $\mathcal{C}_k=\{i\mid y_i=k\}$ be the index set of its
samples, and define the within-class restriction
$E_k = E[\mathcal{C}_k,\mathcal{C}_k]$. We then compute a separate dendrogram
for each class:
\begin{equation}
  \mathcal{T}_k = \mathrm{AHC}_{\mathrm{Linkage}}(E_k),
\end{equation}
where $\mathrm{Linkage}$ denotes the hierarchical linkage operator (Ward's
criterion in our experiments). Importantly, each tree $\mathcal{T}_k$ is formed
only by samples of class $k$, but distances in $E_k$ are computed from
full-context behavior profiles $\mathcal{M}[p,:]$, so within-class organization is
informed by how each sample relates to all classes.

\subsubsection{Step 2: Identifying Coherent Groups and Selecting Clusters}
\label{subsubsec:second_step}

Given the class-wise dendrograms $\{\mathcal{T}_k\}_{k=1}^K$, we seek a flat
partition that exposes the most coherent intra-class substructure induced by the
full-context geometry. Let $\mathcal{P}(\mathcal{T}_k)$ denote the set of
candidate partitions obtainable by cutting $\mathcal{T}_k$ at admissible merge
levels. We select the best partition by maximizing the silhouette coefficient:
\begin{equation}
  \mathcal{G}_k^* = \arg\max_{\mathcal{G} \in \mathcal{P}(\mathcal{T}_k)}
    \mathrm{SilhouetteCoefficient}(\mathcal{T}_k;\mathcal{G}),
  \label{eq:group_of_clusters}
\end{equation}
where $\mathcal{G}_k^*=\{\mathcal{G}_{k,i}^*\}_{i=1}^{m_k}$ is the optimal set of
clusters for class $k$.

\paragraph{Ranking clusters by coherence.}
Let $s_{k,i}$ denote the mean silhouette of cluster $\mathcal{G}_{k,i}^*$ and
define the per-class average
\begin{equation}
  \bar{s}_k = \frac{1}{m_k} \sum_{i=1}^{m_k} s_{k,i}.
  \label{eq:cluster_mean_silhouette}
\end{equation}
Let $\pi_k$ be a permutation of $\{1,\dots,m_k\}$ such that
$s_{k,\pi_k(1)} \ge \dots \ge s_{k,\pi_k(m_k)}$.

\paragraph{Cluster selection via $h(\cdot,\tau)$.}
Given a shared selection parameter $\tau$, we retain a subset of coherent
clusters per class as
\begin{equation}
  \mathcal{G}_k(\tau) = h(\mathcal{G}_k^*,\tau) =
  \begin{cases}
    \{ \mathcal{G}_{k,\pi_k(i)}^* \}_{i=1}^{\tau}, &
      \text{if } \tau \in \mathbb{N} \text{ (top-$\tau$ selection)},\\[4pt]
    \{ \mathcal{G}_{k,i}^* \mid s_{k,i} \ge \bar{s}_k \}, &
      \text{otherwise} \text{ (above-average selection)}.
  \end{cases}
  \label{eq:cluster_selection}
\end{equation}
We then form the global set of selected coherence groups
$G = \bigcup_{k=1}^{K} \mathcal{G}_k(\tau)$ and reindex it as
$G=\{G_j\}_{j=1}^{J}$ to avoid carrying the class/cluster double index
$(k,i)$. From this point onward, classes are not used explicitly in the method
definition, since all operations are performed cluster-wise.

\subsubsection{Step 3: Aggregated-Cluster Embedding and Context Feature Filtering}
\label{subsubsec:third_step}

Step~3 turns the selected coherence groups $G_j$ into an explicit inductive
feature map $\phi_f(\cdot)$ by (i) selecting a small number of inner references
per group, (ii) building reference-conditioned weighted representations of each
group, and (iii) aggregating those representations into scalar coordinates.

For each coherence group $G_j$, define the within-group submatrix
\begin{equation}
  \mathcal{L}_j = \mathcal{M}[G_j,G_j] = \big[m_{x,y}\big]_{x,y\in G_j}.
\end{equation}

\paragraph{Step 3.a: Inner-reference selection per group.}
Let $F\ge 1$ be the desired number of references per group and
$\mathrm{ref\_mode}\in\{\mathrm{close},\mathrm{far}\}$ the selection policy.
We use a centrality functional based on the Euclidean distance between rows of
$\mathcal{L}_j$:
\begin{equation}
  \Delta_j(x,\Theta)
    = \frac{1}{|\Theta\setminus\{x\}|}
      \sum_{x' \in \Theta\setminus\{x\}}
      \Big\lVert \mathcal{L}_j[x,:]-\mathcal{L}_j[x',:]\Big\rVert_2,
  \qquad x\in \Theta\subseteq G_j,
  \label{eq:step3_centroid_score}
\end{equation}
and choose $R_j=\{r_j^i\}_{i=1}^{F}\subseteq G_j$ using either a medoid-like
(\texttt{close}) or a covering (\texttt{far}) strategy (Appendix~\ref{app:ref_selection}).

\paragraph{Step 3.b: Reference-conditioned weighting and dissimilarity vectors.}
For each reference $r_j^i\in R_j$, we compute a within-group weight vector
$\omega_j^i\in\mathbb{R}^{|G_j|}$ by measuring how similar each member of the
group is to the reference in the $\mathcal{L}_j$ row space:
\begin{equation}
  \omega_j^i
    = 1 - \mathrm{norm}_{[0,1]}\Big(
        \delta_{\text{eucl}}\big(\mathcal{L}_j[r_j^i,:],\mathcal{L}_j\big)
      \Big),
  \label{eq:step3_weight}
\end{equation}

\begin{equation}
  \mathrm{norm}_{[0,1]}(v)
    = \frac{v - \min(v)}{\max\{\max(v)-\min(v),1\}},
  \label{eq:norm}
\end{equation}
where $\delta_{\text{eucl}}(v,A)$ returns the vector of Euclidean distances from
$v$ to each row of $A$. We then define, for any object $x$ (training or unseen),
its behavior profile restricted to the group as
$\mathcal{M}[x,G_j]=\{m_{x,y}\mid y\in G_j\}$. The reference-conditioned dissimilarity
vector is defined as
\begin{equation}
  V^i_{j}(x)
    = \delta_{\text{eucl}}\Big(
        \mathcal{M}[x,G_j],\; \mathcal{L}_j \cdot \mathrm{diag}(\omega_j^i)
      \Big)
    \in \mathbb{R}^{|G_j|},
  \label{eq:values}
\end{equation}
which yields one Euclidean dissimilarity per member of $G_j$ between $x$'s
restricted profile and the reference-weighted group representation.

\paragraph{Step 3.c: Aggregation into Scalar Coordinates and Embedding Synthesis.}
Let $f\in\{\min,\mathrm{mean},\max,\mathrm{median},\|\cdot\|_2\}$ be an aggregation
functional. For each $(j,i)$, define the scalar coordinate
$\phi_{j,i}(x)=f(V_{j,i}(x))$. The final embedding is then the concatenation

\begin{equation}
\phi_f(X) = \bigg[\big[f(V^i_{j}(X))\big]_{i=1}^{F}\bigg]_{j=1}^{G} \in \mathbb{R}^{|X|\times (F|G|)},
\label{eq:phi}
\end{equation}

\paragraph{Inductive constraint and context feature filtering.}
At inference time, $\phi_f(x^\ast)$ is computed solely through compression dissimilarity evaluations against the training-derived sets $G_j$. Consequently, it
is not necessary to compute distances to all objects in $X$: only distances to
the union of selected group members $\bigcup_j G_j$ are required, and the size
of this subset is controlled by the Step~2 selection parameter $\tau$.

\begin{sidewaysfigure}
  \centering
  \includesvg[width=0.875\textwidth, inkscapearea=drawing, inkscapelatex=false]{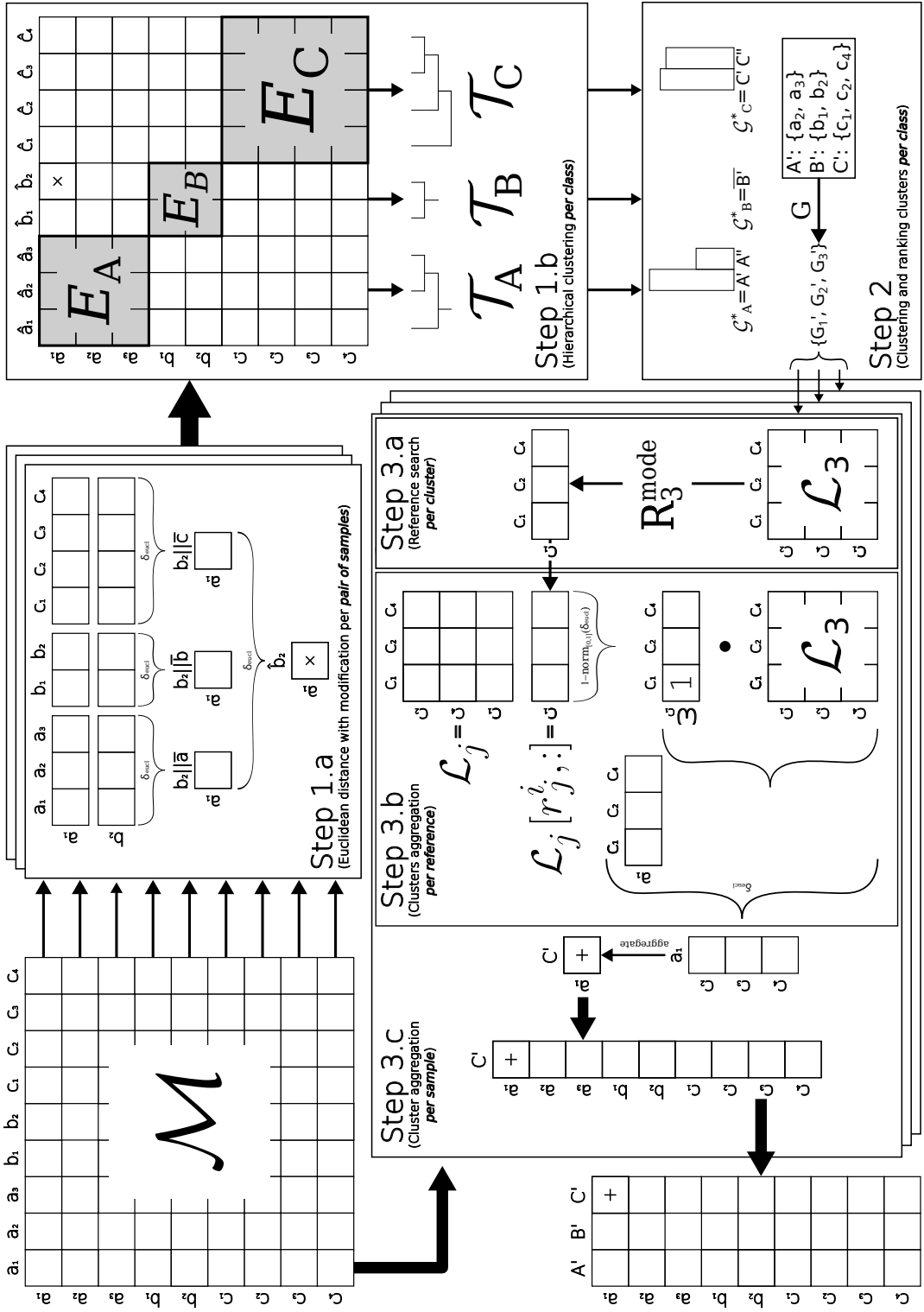}
  \caption{\textbf{Overview of the proposed method.} In this example, rows and columns are labeled as $a_i$, $b_i$, or $c_i$ to reflect class membership. The process starts from the
    full compression-dissimilarity matrix $\mathcal{M}$ (top-left). In \textbf{Step~1.a}, rows
    of $\mathcal{M}$ are interpreted as behavior profiles and compared via an Euclidean
    layer (optionally using the blockwise refinement of
    Section~\ref{subsubsec:add_step}) to obtain the behavior-similarity matrix
    $E$. In \textbf{Step~1.b}, linkage is performed \emph{separately} within each
    class using the class-restricted matrices $E_k$, producing one dendrogram
    $\mathcal{T}_k$ per class. In \textbf{Step~2}, each $\mathcal{T}_k$ is cut to
    obtain a partition that maximizes the silhouette coefficient; clusters are
    ranked by mean silhouette and the most coherent ones are retained, yielding
    a global set of coherence groups $G=\{G_j\}$. In \textbf{Step~3.a}, a small
    set of inner references $R_j$ is selected per group (either central or
    covering). In \textbf{Step~3.b}, for each selected reference, a weight vector
    $\omega_j^i$ is computed to emphasize group members aligned with that
    reference, and a reference-conditioned dissimilarity vector $V_{j,i}(x)$ is
    obtained for each sample $x$. In \textbf{Step~3.c}, each $V_{j,i}(x)$ is
    aggregated into a scalar coordinate and concatenated across groups and
    references to form the inductive embedding $\phi_f(x)\in\mathbb{R}^{F|G|}$.}
  \label{fig:method_overview}
\end{sidewaysfigure}

\subsubsection{Optional refinement: Blockwise Euclidean distance over class feature blocks}
\label{subsubsec:add_step}

In some scenarios (notably when class blocks in $\mathcal{M}$ have different magnitudes or
effective sizes), the standard Euclidean layer in (\ref{eq:euclidean}) can be
dominated by a subset of features. To mitigate this effect, we define a
class-block-aware alternative that sums per-class $\ell_2$ distances:
\begin{equation}
  \delta_{\text{alt\_eucl}}(p, q)
    = \sum_{k=1}^{K}
        \Big\lVert \mathcal{M}[p,\mathcal{C}_k]-\mathcal{M}[q,\mathcal{C}_k]\Big\rVert_2
    = \sum_{k=1}^{K}
        \left(
          \sqrt{\sum_{x \in \mathcal{C}_k} (m_{p, x}  - m_{q, x})^2}
        \right),
  \label{eq:our_euclidean}
\end{equation}
which can be interpreted as an $\ell_1$ aggregation of per-block $\ell_2$
distances. In our experiments, using $\delta_{\text{alt\_eucl}}$ improved the
stability of the Euclidean layer in settings where class-wise feature blocks are
unbalanced (e.g., fragment-based datasets).

\section{Materials and Methods}
\label{sec:materials_methods}

This section describes the materials and methodological components used to
evaluate our approach. We first present the datasets employed in the
experiments, which were selected to span multiple levels of complexity and
represent both textual and audio-based modalities. We then detail the relative
compressor used to compute similarity (RLZAP), and discuss the necessary
adaptations made over the final NRCs for heterogeneous data types.

\subsection{Datasets}
\label{subsec:materials}
To evaluate the proposed methodology, we assembled three datasets of varying
complexity, categorized as \textit{easy}, \textit{medium}, and \textit{very
  hard}. These datasets encompass text- and audio-based samples, enabling a
comprehensive assessment of our approach across distinct data modalities and
classification challenges. A brief summary of each dataset is included in
Table~\ref{tab:datasets_summary}.

\begin{table} 
  \caption{\textbf{Summary of datasets used in the experiments}. In all experiments
    using a K-fold setup, $K=5$ is applied for training and testing
    classification.}
  \label{tab:datasets_summary}
  \centering
  \begin{tabular}{lccccc}
    \toprule
    \textbf{Dataset} & \textbf{Modality} & \textbf{\#Classes} & \textbf{\#Samples per Class} & \textbf{\#Total} & \textbf{Formats} \\
    \midrule
    Easy       & Text        & 2 authors   & 67–71 & 138 & 1 \\
    Medium     & Text        & 6 authors            & 40 & 240 & 1 \\
    Very hard  & Audio       & 5 species            & 40 recordings & 2817 (fragments) & 5 \\
    \bottomrule
  \end{tabular}
\end{table}

\paragraph{Easy dataset} This dataset comprises short stories authored by
H.~P.~Lovecraft and Edgar Allan Poe, sourced from open repositories (Lovecraft
GitHub corpus~\cite{smithVilmibmLovecraftcorpus2024} and Poe Kaggle
corpus\footnote{The authors mention that they assembled the dataset from Project
  Gutenberg.}~\cite{EAPoesCorpus}). It includes 67 Lovecraft and 71 Poe stories,
all in plain text and in English. The dataset has not undergone preprocessing.
Its classification as \textit{easy} stems from its simplicity: only two
well-separated classes with sufficient samples per class, which facilitates
straightforward clustering and context evaluation.

\paragraph{Medium dataset} This dataset increases complexity by incorporating
short stories from six authors: William Sydney Porter (O.~Henry), Hans Christian
Andersen, Anton Pavlovich Chekhov, Nathaniel Hawthorne, H.~P.~Lovecraft, and
Edgar Allan Poe. Using Project Gutenberg’s open library~\cite{ProjectGutenberg},
we selected 40 short stories per author, all in English and plain-text format.
This balanced but more diverse dataset introduces additional intra- and
inter-class variability, meriting its classification as \textit{medium}
complexity.

\paragraph{Very hard dataset} This dataset shifts focus to audio samples,
assembled from the Xeno-Canto database~\cite{XenocantoBirdSounds2024}, an open
repository of bird (and some insect) recordings submitted by the public. As a
result, the dataset is not curated, introducing considerable variability in
audio quality, duration, and labeling accuracy. This dataset comprises
recordings from five species: one grasshopper (\textit{Chorthippus biguttulus})
and four birds (\textit{Anas platyrhynchos}—mallard, \textit{Anhima
  cornuta}—horned screamer, \textit{Anser albifrons}—greater white-fronted
goose, and \textit{Anser anser}—greylag goose). The species were selected to
span varying taxonomic distances, with three birds from the same family and two
sharing the same genus. Each species contributes 40 audio recordings, which were
standardized by subsampling to 16~kHz and compressing to mono-channel MP3 files
at 40~kbps. To enable cross-modal compression analysis, each recording was
segmented into consecutive 2-second fragments
(following~\cite{sarasaAutomaticTreatmentBird2018}); end of audio segments were
mostly discarded\footnote{As not all audios have an even duration, the last
  segment sometimes tends to be unusable.}. The remaining segments were
converted to WAV and text-based formats. The text representation was generated
by encoding the audio frequencies (on a logarithmic space) sorted by power, as
textual strings. This combination of taxonomic proximity, open recording
conditions, and heterogeneous formats makes the dataset a demanding benchmark
for compression-based similarity.

In addition to its taxonomic diversity and heterogeneous format structure, the
inherent openness of the Xeno-Canto dataset introduces severe challenges rarely
found in controlled benchmarks. Audio samples are collected and labeled by
non-expert contributors under varying recording conditions, leading to
inconsistencies in both label accuracy and acoustic content. As a result, a
labeled sample may or may not contain the vocalization of the target species,
and may additionally include sounds from other birds, insects, humans, or
environmental noise. Furthermore, each species exhibits considerable intra-class
variability: individual recordings may capture songs, calls, mechanical noises
(e.g., wing flaps, pecking), or overlapping behaviors. Compounding these
difficulties, the use of compressed and segmented audio—later encoded into
textual representations—introduces further distortions that must be resolved
implicitly by the compression-based similarity. Crucially, no denoising,
filtering, or preprocessing was applied beyond standardization and
fragmentation, in order to preserve a realistic evaluation setting. Taken
together, these factors—label noise, representation shift, inter- and
intra-class variability, open-world contamination, and lossy format
conversion—render this experiment orders of magnitude more complex than the
previous ones, both in terms of data quality and methodological requirements.

\subsection{RLZAP}
\label{subsec:rlzap}

In this work, the relative compressor used to measure Normalized Relative
Compression  (NRCs) is based on the rlzap compressor, as proposed in
\cite{coxRLZAPRelativeLempelZiv2016}\footnote{The code is available in the
  original authors repository: \url{github.com/farruggia/rlzap}}. Rlzap builds
upon the general rlz (Relative Lempel-Ziv) framework, initially introduced in
\cite{zivMeasureRelativeEntropy1993}, and further refined in subsequent works
\cite{kuruppuRelativeLempelZivCompression2010b,ferradaRelativeLempelZivConstantTime2014,doFastRelativeLempelZiv2014}.
The core idea of rlz-based compressors is to encode a target sequence by
referencing substrings from a fixed reference object, using a sequence of
position–length pairs. This allows a notion of similarity to emerge between
target and reference, based on the ability to reproduce the former using
substrings of the latter.
 rlzap introduces an enhancement to this basic scheme through the use of adaptive
pointers, which allow for small local discrepancies in the matching process
(such as insertions, deletions, or multi-character substitutions) to be handled
more gracefully. While the original motivation for rlzap lies in the compression
of genomic data, which typically exhibits high local similarity and structural
shifts, this same property can result in suboptimal behavior when applied to
more general or heterogeneous data. Nevertheless, this specificity is not
inherently problematic in our context. On the contrary, it is desirable: our
method deliberately seeks compressors that impose stricter matching constraints,
as the goal is not to find broad statistical similarity but to identify highly
discriminative patterns that reflect the context selected for our analysis.

The selection of rlzap was also driven by its availability as an open-source
tool and its operational simplicity, which facilitated integration in our
pipeline. Our intention was not to optimize compression ratio or explore a broad
spectrum of compressors, but to establish a representative, robust scheme with
consistent behavior across samples. As a member of the rlz family, rlzap
satisfies this criterion while also enabling a fine-grained control over the way
differences between objects are captured.

As introduced in Section~\ref{subsec:compression}, a common issue in all
compression-based dissimilarity measures is the disproportionate influence that large
discrepancies in object size exert in compression dissimilarities. Whether measuring
the similarity between two objects (as in NCD) or between an object and a fixed
reference (as in NRC), large discrepancies in size or structure can cause
compression artifacts that distort the resulting distances. In the case of the
NRC, such effects are particularly amplified, as it relies completely in the
ability of $C(x \Vert y)$ to handle object's size differences. The asymmetry of
using a the NRC as a base measure, exposes the full weight of the reference's
encoding (see equation \ref{eq:nrc}); whereas NCD partially compensates via its
use of $\min(C(x), C(y))$ and $\max(C(x), C(y))$ in the denominator.

To mitigate the object-specific biases, we apply a row-wise standardization to
the resulting dissimilarity matrix, normalizing each row using its mean and standard
deviation. This reduces the impact of dominant trends in each object’s distance
profile while preserving finer-grained differences that are more relevant to
downstream tasks such as clustering or classification. This process is performed
after conforming the dissimilarity matrix. When possible, normalization parameters
are computed from an external dataset that serves solely as a statistical
baseline\footnote{This step is included as an additional mechanism to ensure
  that we are not introducing any bias in the row-wise standardization process. Hence,
  when possible, both scenarios are included in our results.} remains
excluded from all training and testing procedures; in this case, means and
standard deviations are estimated from the distances between the original
objects and the external set. If another source is unavailable, the available
data is used instead, deriving the statistics from the training set (of each
KFold, described in detail in Section~\ref{sec:experiments}). This
approach builds on the empirical observation that each object tends to produce a
stable distance signature when compared to a fixed set of references—consistent
in overall shape but subject to scaling. As a result, row-wise standardization
improves comparability while supporting incremental extension, since new objects
can be standardized using precomputed statistics without requiring full
recomputation.

During experiments, we observed that rlzap, being primarily designed for text
and genomic data, fails or crashes in some cases when applied to files in binary
formats, such as mp3 or wav files. To ensure compatibility, all non-textual
files were converted into ASCII-based hexadecimal representations. As detailed
in Section \ref{sec:experiments}, we considered three file formats for the
Xeno-canto dataset: mp3, wav, and plain text. The mp3 and wav files were
transformed into textual form by interpreting their binary content as
hexadecimal strings, resulting in the variants hex\_mp3 and hex\_wav. This
preprocessing step ensured that all input data could be processed uniformly by rlzap, enabling consistent evaluation of similarity across heterogeneous file
types.
\section{Experiments}
\label{sec:experiments}

In this section we evaluate our approach on three datasets of increasing
difficulty, ranging from a relatively simple binary text authorship problem to a
heterogeneous multi-class audio scenario. For each setting, we consider both
classification and clustering as complementary validation tools: classification
performance on held-out test data is used as our primary metric, while
clustering analyses on the induced embeddings provide an additional diagnostic
view of the class structure captured by the method. Across the three datasets,
we are particularly interested in how performance degrades as task complexity
increases, and in how our approach compares to compression-based baselines and
straightforward knn schemes built directly on the original compression dissimilarities.

We designed our experimental procedure to systematically evaluate how each
method performs under varying conditions and datasets. Specifically, our aim was
to isolate the contribution of each processing step, ensuring that any observed
differences in performance could be attributed to the methods themselves rather
than external confounding factors. The procedure is structured as follows:
\begin{enumerate}
\item \textbf{K-Folds:} We employ a standard $K$-fold cross-validation procedure
  to ensure robust performance estimates. Additionally, because our input takes
  the form of a square distance matrix (where each column corresponds to one of
  the rows), it is critical to remove from each fold any columns
  associated with test samples. Otherwise, the training observations would
  indirectly ``see'' test samples as features (through the compression
  distances), leading to data leakage. Hence, for each fold, we only keep those
  columns corresponding to objects in the current training set. For all of our
  experiments, we used $K = 5$.

\item \textbf{Object Analysis:} In this step, we apply different methods to
  investigate their respective impacts on the results. This is where our
  approach and the alternative procedures each differ for every experiment, as
  they execute different methods to split classes into new clusters. The output
  of this step is a pair of arrays containing (1) the cluster label of each
  observation and (2) the reference indices chosen by the method. The first
  array represents the ``contextual subsets'' identified by the algorithm, while
  the second array denotes the selected references for each subset. Following
  the methodology described in Section~\ref{sec:methodology}, for our approach
  this phase includes Step~1 and Step~2 described in
  Sections~\ref{subsubsec:first_step} and \ref{subsubsec:second_step} (Steps~1.a--1.b and Step~2 of Figure~\ref{fig:method_overview}).

\item \textbf{Feature Computation:} Next, we aggregate clusters, reference
  objects, and the complete set of observations to compute the final features.
  Although this aggregation process is part of our proposed method, we separate
  it as an independent stage to isolate the performance contribution of our
  methodology. This approach allows a clearer assessment of how our method
  influences the resultant feature space. For our approach, this part applies
  the step three of our methodology described in
  Section~\ref{subsubsec:third_step} (Steps~3.a--3.c of
  Figure~\ref{fig:method_overview}).

\item \textbf{Performance Scoring:} Finally, we train a classical
  classification model (random forest) on the newly generated features and
  measure its $F_1$ score on both training and test folds, and average them.
  We also compute the silhouette coefficient as an auxiliary metric to assess
  the quality of the transformed data. In experiments involving segmented
  samples (\textit{e.g.} audio segments), we select the segment with the highest
  prediction probability from the model as the representative reference
  (\textit{i.e.} the one that decides the class for the whole file) for the
  original file.
\end{enumerate}

\subsection{Experimental Setup and Comparative Analysis}

As our method essentially has two main stages: finding clusters and their references,
and then aggregating data into new features, we have employed various methods to
provide a comprehensive evaluation of our approach. To evaluate the entire
approach from an analogous perspective, we have used a combination of different
feature selection methods paired with Kmeans. Furthermore, to determine the
significance of the first step (\textit{i.e.} finding relevant clusters), we
include two different random methods together with our feature aggregation
methodology. And finally, to analyze the impact of the original compression dissimilarities, we utilize a K-nearest neighbors (Knn) classifier, which bypasses the
selection entirely and uses the complete distance matrix as input. Each of these
methods is detailed as follows:

\begin{itemize}
\item \textbf{Random:} Both clusters and references are selected randomly from
  each class. To account for variability, we average the score metrics over
  multiple iterations.

\item \textbf{Dummy:} This setup is similar to the \emph{Random} method, but
  references are selected from the entire dataset instead of being restricted by
  class, while each class remains its own cluster. Again, we average the score
  metrics over multiple iterations.

\item \textbf{KBest/KMeans Clustering (from now on ``kbest''):} We utilize
  several classical feature-selection techniques (mutual information,
  chi-square, and ANOVA) to identify the best features based on the training
  class labels. We then run a standard KMeans algorithm, setting the number of
  clusters equal to the number of selected features. This yields a set of
  clusters aligned with the ``best'' features according to these selection
  methods.

\item \textbf{Knn:} We also implement a knn classifier, as many other works in
  the literature use it as a first approach to NCD classification. For this
  method, the steps 2 and 3 (Sections \ref{subsubsec:second_step} and
  \ref{subsubsec:third_step}) are skipped as it produced better results on top
  of being closer with its use in the literature. Additionally, knn always
  relies on measuring how ``close'' or ``similar'' data points are to each
  other. While it's possible to provide these ``distances'' to the model
  beforehand (as precomputed distances), we've opted to let the model calculate
  them itself using Euclidean distances as we have observed that this approach
  yields better results.

\item \textbf{Our Approach: Context \textit{Steering}} We implement the process
  detailed in this paper, described in depth in Section~\ref{sec:methodology}.
\end{itemize}

In the process of assembling the embedding, there are parameters that can be
optimized for different scenarios. However, it is important to pay special
attention to those parameters that modify the size of the final transformation.
These are the number of clusters and number of references per cluster. We have
observed that our approach tends to optimize far better than the alternatives,
that tend to always improve as more elements are included in the final
embedding. This represents a crucial element when applied in real world
scenarios, as the number of objects required for the embedding not only alters
accuracy but also limits applicability. This is specially challenging for Knn,
which uses every element in the set, and hence, has a prohibitive size for most
cases. As each new observation requires the recomputation of every compression
dissimilarity against every object of the set (for precomputed distances, if an
additional distance measure is used, it also require to recompute the distances
against each feature vector).

In our experiments, we have taken into account this element in the process of
choosing the number of final objects for each experiment and method. Each
experimental setup and results format, are described in their respective
section.

\subsection{Easy Dataset: Short Stories of H.P. Lovecraft and Edgar Allan Poe}
\label{subsec:easy_exp}

In this experiment, we aim to verify whether our pipeline and all the chosen
compressors can effectively handle a relatively simple scenario, thereby
confirming that the validation procedure yields reasonable results. Essentially,
this serves as a preliminary validation to ensure that the approach is reliable
before applying it to more complex cases.

We begin by examining the degree of class separability provided by four
different compressors (from now on, the starting point to every
experiment)--three using NCD formulation: zlib, lzma and bz2; and one with NRC: rlzap--using short stories by H.P.~Lovecraft and Edgar Allan Poe (described in
detail in Section~\ref{subsec:materials}). From this point forward, the rlzap
results are presented in three variants:
\begin{itemize}
    \item Rlzap, which shows the NRC distances without any additional processing.
    \item Rlzap standardized, which applies the row-wise
      standardization discussed in Section~\ref{subsec:compression}. This
      category is further divided in external std. and pipeline
      std., as introduced in Section \ref{subsec:rlzap}. The former uses a
    standardization with mean and standard deviation from an external dataset,
    while the latter is standardized inside of the K-Folds, using statistical
    values only from the training set (for each fold respectively).
\end{itemize}

Figure~\ref{fig:hctrees_poecraft} illustrates the initial clustering outcomes
under AHC for each compressor. While zlib, lzma, and
bz2 do not display a perfectly accurate separation of the two classes,
each nonetheless reveals distinct subgroups that appear to align reasonably well
with the underlying class structure. Conversely, rlzap shows
little to no class grouping, whereas rlzap (standardized) (with row-wise
standardization) exhibits almost perfect class separability.

It is important to note that, for this experiment, the mean and standard
deviation used in the row-wise standardization step for rlzap were
calculated from the current sample set itself, rather than from an external
dataset. Owing to the limited number of short stories available, it was not
feasible to hold out a larger secondary sample exclusively for computing
normalization statistics from these two authors alone. Nevertheless, for the
main validation process, this standardization is always computed using only
training samples.

\begin{table} 
  \label{tab:easy_dataset_scoring_results}
  \caption{\textbf{Classification results for our context \textit{steering}
      approach}: Comparison of performance metrics for each compressor on the
    \emph{Easy} dataset, measured via average test and training F1 scores and
    silhouette coefficients. Identical parameters were applied across runs,
    selected from a broader pool explored for different compressors and
    configurations. Since all compressors achieved at least a 0.9 average F1
    score on the test set, further parameter optimization was not pursued,
    aligning with the limited scope of this experiment. The Silhouette
    coefficients for the training splits do not exactly match the results shown
    in Figure~\ref{fig:hctrees_poecraft}. This discrepancy exists because the
    experiment in the figure used all objects without a prior train-test split.
    However, the relative performance between compressors remains similar in
    both scenarios.
  }

  \centering
  \begin{tabular}{lrrrr}
    \toprule
    \textbf{Compressor} & \textbf{Test F1 Score } & \textbf{Train F1 Score } & \textbf{Test silhouette } & \textbf{Train silhouette }\\
    \midrule
    Bz2                   & 0.941     & 0.996     & 0.201     & 0.211     \\
    Lzma                  & 0.956     & 0.993     & 0.120     & 0.120     \\
    Rlzap (unstandardized)       & 0.663     & 0.947     & 0.017     & 0.013     \\
    Zlib                  & 0.934     & 0.991     & 0.162     & 0.163     \\
    Rlzap (pipeline std.) & 0.905     & 0.995     & 0.440     & 0.467     \\
    \bottomrule
  \end{tabular}
\end{table}

Table~\ref{tab:easy_dataset_scoring_results} summarizes the performance metrics
for each compressor, showing that all of them--except rlzap (unstandardized)--report satisfactory results. This finding is consistent
with our expectations, given the relatively low complexity of the dataset.
Another noteworthy observation is that the silhouette coefficient for rlzap (pipeline std.) is almost double that of the best-performing conventional
compressors, which may indicate a genuinely stronger performance for rlzap (pipeline std.) or, alternatively, a side effect of the row-wise standardization
in a two-class scenario. It is important to notice that the silhouette
coefficients are similar but not equal to the ones produced in
Figure~\ref{fig:hctrees_poecraft}. The reason behind this is that, in the
training step, the number of samples is one fifth smaller, given that the K-folds
step uses five folds. And thereby, the procedure of assembling the tree is not
quite the same.

\begin{figure}[!ht]
  \centering
  \includesvg[width=\textwidth]{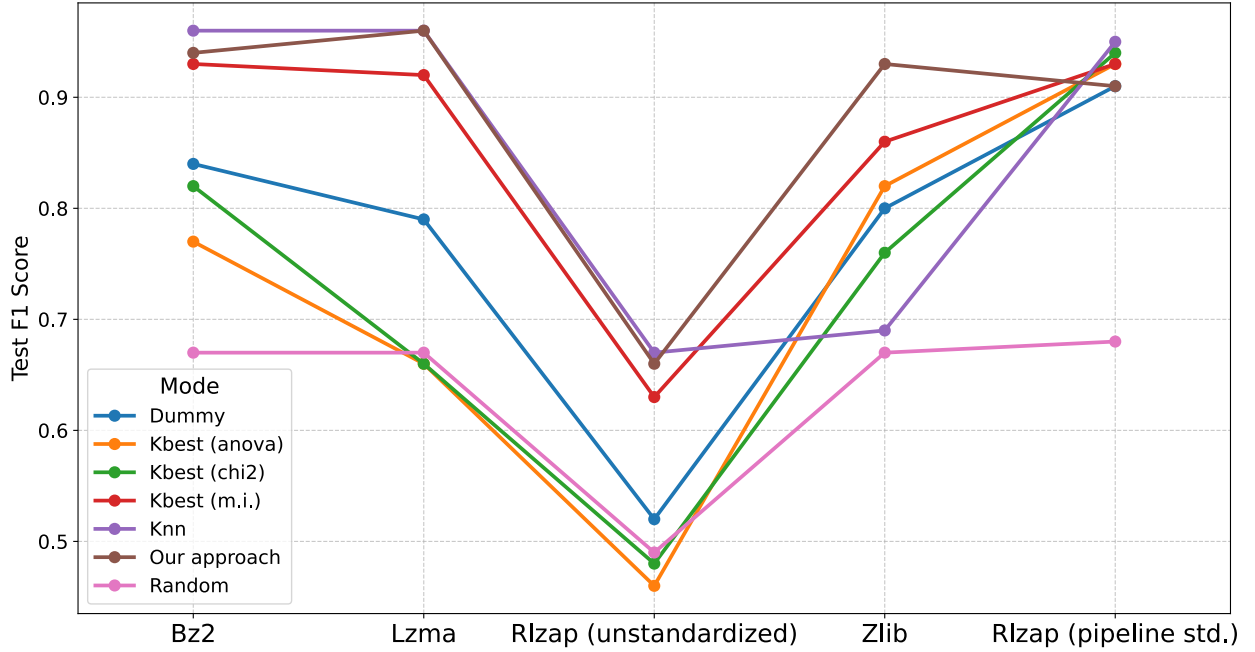}
  \caption{ \textbf{Test F1 scores for various compressors and methods} under the
    validation procedure. Each compressor uses the same number of features
    selected by our approach for that compressor, ensuring a fair comparison
    with alternative methods: \emph{kbest (anova)}, \emph{kbest (chi2)},
    \emph{kbest (m.i.)}, \emph{knn}, \emph{dummy}, and \emph{random}. Parameter
    selection for our approach was performed using a pool of multiple runs
    across different compressors and configurations. Since the purpose of this
    experiment is to stablish a baseline for every scenario, no additional
    optimization was conducted. }
  \label{fig:f1_test_score_plot}
\end{figure}

To illustrate these comparisons more clearly,
Figure~\ref{fig:f1_test_score_plot} depicts the \emph{Test F1 scores} for each
combination of compressor and method. Although \emph{kbest} can sometimes match
or even outperform our approach--especially in rlzap--this is not entirely
unexpected, given the differing objectives of each algorithm. These results
confirm that for an \textit{easy} dataset, our pipeline is robust and capable of
delivering strong baseline performance, paving the way for more complex
scenarios. Finally, each mode (\emph{dummy}, \emph{random} and \emph{kbest
  anova, chi2} and mutual information (\emph{m.i.})) uses the same number of
features that our approach selected for each respective compressor. For the
remaining experiments on this work, these alternative modes will work without
any feature limitation, in order to properly compare them against our approach.

\subsection{Medium Dataset: Short Stories of Six Different Authors}
\label{subsec:med_exp}

Following the encouraging results from the \emph{Easy} dataset, where
compressors could readily differentiate between two authors (even under
\emph{random}), we now shift to a more complex, multiclass scenario. In
particular, we expand from two to six classes by adding four additional authors
(this dataset is described in depth in Section~\ref{subsec:materials}).
For each class, we randomly select 40 samples to mitigate potential class
imbalance and ease the burden on the final prediction model.

We systematically compare our approach against alternative methods by conducting
a parameter grid search for each compressor, identifying the best scores for
every configuration. As we point out before, the number of samples used in the
embedding is greater for the alternative methods than for our approach, as in
similar conditions, the alternatives produce artificially lower results.

The results of these best runs appear in Table~\ref{tab:medium_gridresults}.
Despite all compressors exhibiting acceptable performance--including rlzap
(unstandardized)--the silhouette coefficient reveals that only rlzap and bz2 produce
``reasonable'' cluster structures. The near-zero silhouette values suggest
each class may be split into smaller ``subclasses'' that remain distinguishable
from other classes. For simplicity, we have removed the rlzap (unstandardized)
compressor in the folowing experiments, as including an additional compressor
would difficult the readability of each figure.

\begin{table} 
  \caption{ \textbf{Medium dataset scoring results.} Each entry corresponds to
    the best run found for that compressor from a grid search over multiple
    parameters. \textit{Rlzap (external std.)} and \textit{Rlzap (pipeline std.)} both have
    row-wise standardizations, being the first one performed using external
    statistical values, and the second one using the ones from each training
    subset. \textit{Rlzap (unstandardized)} represents the unstandardized distances without any normalization.
    All compressors demonstrate a reasonable F1 score on the test set with our
    approach, although \textit{Rlzap (unstandardized)} remains the weakest performer, as
    expected. }
  \label{tab:medium_gridresults}
  \centering

  \begin{tabular}{lrrr}
    \toprule
    \textbf{Name} & \textbf{Test F1 score} &  \textbf{Silhouette (Test clusters)} & \textbf{Silhouette (Train clusters)}\\
    \midrule
    Bz2 & 0.9256 & 0.0327 & 0.0711 \\
    Lzma & 0.8806 & -0.1072 & -0.0281 \\
    Zlib & 0.8732 & -0.0470 & 0.0169 \\
    Rlzap (external std.) & 0.8802 & 0.0608 & 0.0903 \\
    Rlzap (unstandardized) & 0.7772 & -0.1400 & -0.0783 \\
    Rlzap (pipeline std.) & 0.8506 & 0.1093 & 0.1523 \\
    \bottomrule
  \end{tabular}
\end{table}

In these experiments, the silhouette coefficient must be interpreted carefully,
as it only measures the separability of each class from its nearest neighbor. In
the test scenario, additional caution is necessary because each class contains
only eight samples (as detailed in Table~\ref{tab:datasets_summary}, totaling 40
samples per experiment, distributed in $K = 5$ folds). Consequently, the
silhouette coefficient primarily serves as a complement to the F1 metric rather
than as a standalone measure.

It is important to note that, in Table~\ref{tab:medium_gridresults}, the
silhouette score assesses how clearly the clustering of training and test
subsets emerges internally with respect to the classification objective, rather
than evaluating the classifier's performance directly. Specifically, the
similarity of silhouette values between training and test subsets indicates
consistency in how well the proposed methodology organizes the data partitions,
independently of classifier training quality. Given this consistency, we will
report only the silhouette scores corresponding to the test subsets in
subsequent experiments for simplicity. Nonetheless, the silhouette values
presented confirm the increased difficulty of this multiclass scenario compared
to the simpler \emph{easy} dataset.

To gain further insight, we extended our validation by analyzing all possible
subsets of the six classes, ranging from pairs to groups of five. This
comprehensive evaluation bridges the gap between the previous experiment—limited
to only two authors—and the current, more complex scenario involving six
authors. Table~\ref{tab:medium_medians} summarizes the \emph{median} F1 scores
and silhouette coefficients computed across these subsets. Among the compressors
evaluated, bz2 and rlzap particularly stand out, consistently
achieving the highest median F1 scores and notably superior silhouette
coefficients.

\begin{table} 
  \caption{ \textbf{Median F1 and silhouette scores for each compressor} across all
    class combinations (from pairs until groups of five). Bz2 and rlzap (pipline std.)
    achieve notably higher F1 and silhouette values, reflecting improved
    separability under more complex multiclass setups. }
  \label{tab:medium_medians}
  \centering

  \begin{tabular}{lrr}
    \toprule
    \textbf{Compressor} & \textbf{Test F1 (median)} & \textbf{Test silhouette (median)} \\
    \midrule
    Bz2           & 0.925 & 0.158 \\
    Lzma          & 0.879 & 0.025 \\
    Rlzap (pipeline std.)    & 0.809 & 0.006 \\
    Rlzap (external std.) & 0.916 & 0.189 \\
    Zlib          & 0.898 & 0.080 \\
    \bottomrule
  \end{tabular}
\end{table}

\begin{sidewaysfigure}
  \centering
  \includesvg[width=\textwidth]{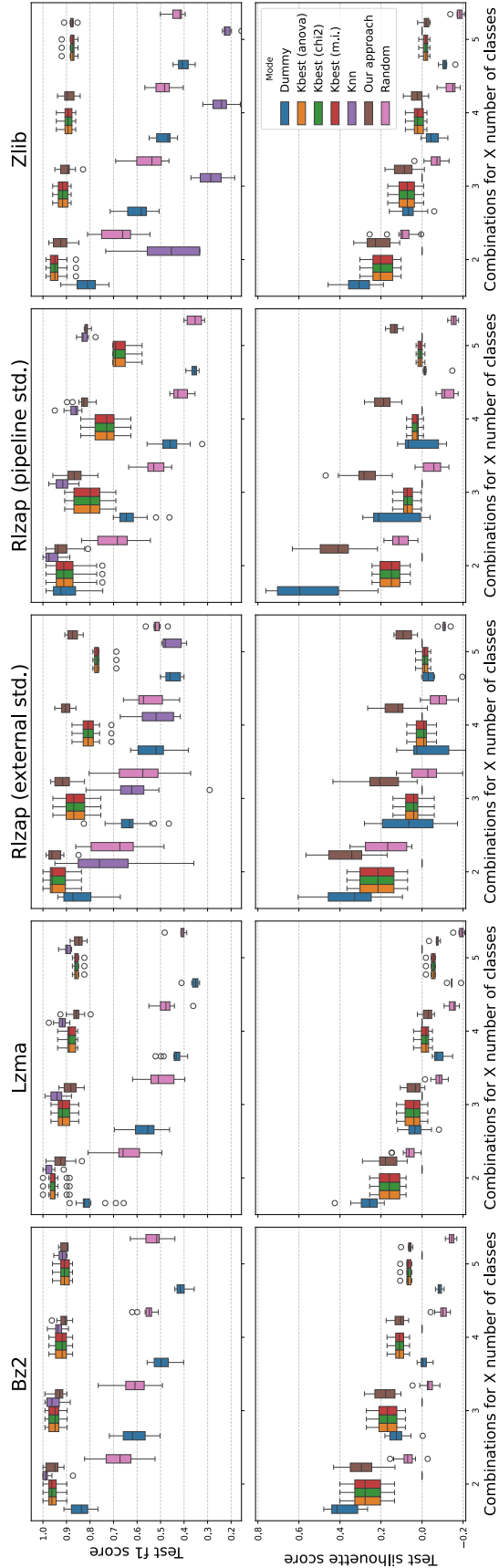}
  \caption{ \textbf{Medium dataset result distributions} for each combination of
    classes during the validation process. The first and second rows display
    test F1 and silhouette scores, respectively. The modes named kbest refer to
    the feature-based alternatives to our approach, while dummy and random modes
    represent the baseline methods described previously. Knn represents a
    Nearest Neighbors computing the Euclidean distance over the original
    dissimilarity matrix. Parameters for each compressor under ``our approach'' come
    from the grid search mentioned earlier, and the sample sizes used in both
    the alternatives and random methods were determined via a simple iteration
    over each candidate pool. It is important to point out that, while our
    approach uses a variable number of elements for each scenario (depending on
    the optimal parameter configuration found for each case), the alternative
    methods use bigger number of objects to compute the embedding. The case of
    knn is particular, as it uses all the objects of the (training) set. }
  \label{fig:boxplot_medium_num_classes}
\end{sidewaysfigure}

Then, we compare how our approach and the alternative methods scale as the
number of classes increases. Figure~\ref{fig:boxplot_medium_num_classes}
illustrates the distribution of F1 and silhouette scores for these two
compressors under different class counts (\textit{i.e.}, isolating and testing
each combination of classes). While bz2 shows a more pronounced improvement in
F1 than in silhouette, rlzap excels in both cases when paired with our approach.
knn, however, presents an interesting behavior. On the one hand, it often
yields F1 scores comparable to those of our method and, in a few configurations,
even slightly higher values (with differences no bigger than 0.07). On the other
hand, there are also configurations in which knn collapses: it fails to produce
reasonable results for zlib and rlzap (under external standardization), with
either very wide F1 distributions or big median scores differences (in some
cases up to 0.65).

Altogether, this highlights both the dependency of knn on a
particularly well-behaved distance geometry and the capability of our method to
remain stable across compressors and class combinations. Finally, we do not
report separate silhouette distributions for knn, since it does not modify the
original dissimilarity matrix and therefore cannot improve the baseline clustering
structure induced by the compressors.

\subsection{Very hard dataset}
\label{subsec:very_hard_exp}

\begin{table}
  \caption{\textbf{Very hard dataset} scoring results. Each entry corresponds to
    the best run found for that compressor from a grid search over multiple
    parameters. Rlzap (external std.) and rlzap (pipeline std.) both include
    row-wise standardizations, the first using external statistics and the
    second using statistics derived from each training subset. Overall, all
    methods experience a considerable drop in F1 score compared to previous
    datasets, reflecting the increased complexity of the task. Finally, the
    silhouette coefficient is really low, highlighting the heterogeneity of the
    dataset (as each class has multiple dominant features to identify, that at
    the same time may or may not share any similarity between them).}
    \label{tab:very_hard_results}
  \centering

  \begin{tabular}{lllrr}
    \toprule
    \textbf{Compressor} & \textbf{Format} & \textbf{Test F1 Score} & \textbf{Test silhouette Score} \\
    \midrule
    Zlib & mp3 & 0.453 & -0.025 \\
     & wav & 0.365 & -0.088 \\
     & hex\_mp3 & 0.517 & -0.038 \\
     & hex\_wav & 0.349 & -0.103 \\
     & txt & 0.772 & -0.039 \\
    \midrule
    Lzma & mp3 & 0.507 & -0.025 \\
     & wav & 0.414 & -0.134 \\
     & hex\_mp3 & 0.357 & -0.046 \\
     & hex\_wav & 0.487 & -0.104 \\
     & txt & 0.568 & -0.076 \\
    \midrule
    Bz2 & mp3 & 0.491 & -0.125 \\
     & wav & 0.517 & -0.084 \\
     & hex\_mp3 & 0.559 & -0.118 \\
     & hex\_wav & 0.448 & -0.074 \\
     & txt & 0.400 & -0.119 \\
    \midrule
    Rlzap (pipeline std.) & hex\_mp3 & 0.493 & -0.030 \\
     & hex\_wav & 0.358 & -0.104 \\
     & txt & 0.694 & -0.096 \\
    \midrule
    Rlzap (external std.)  & hex\_mp3 & 0.477 & -0.297 \\
     & hex\_wav & 0.364 & -0.443 \\
     & txt & 0.669 & -0.077 \\
    \bottomrule
  \end{tabular}
\end{table}
In this final experimental setup, we test the robustness of our approach under a significantly more challenging setting (detailed in Table~\ref{tab:datasets_summary}), designed to emulate real-world conditions. The dataset used in this scenario consists of unprocessed audio samples contributed by users, with minimal curation and a high degree of heterogeneity. Each sample may or may not include vocalizations from the labeled bird species, and may contain additional background sounds or other species, including human voices. Furthermore, the intra-class variability is substantial, due to the diversity of bird vocalizations and incidental acoustic events.

Unlike the previous experiments, which operated on text-based data (see
Section~\ref{subsec:med_exp} and \ref{subsec:easy_exp}), the representation here
is composed by audios. Although these files have been uniformly subsampled
and fragmented into equal-length segments (as described in
Section~\ref{subsec:materials}), no further preprocessing or feature extraction
is applied. Consequently, the solution must account not only
for class-specific patterns, but also for the variability introduced by labeling
uncertainty, noise, inter- and intra-species differences, and heterogeneous
input formats.

\begin{sidewaysfigure}
  \centering
  \includesvg[width=0.8\textwidth]{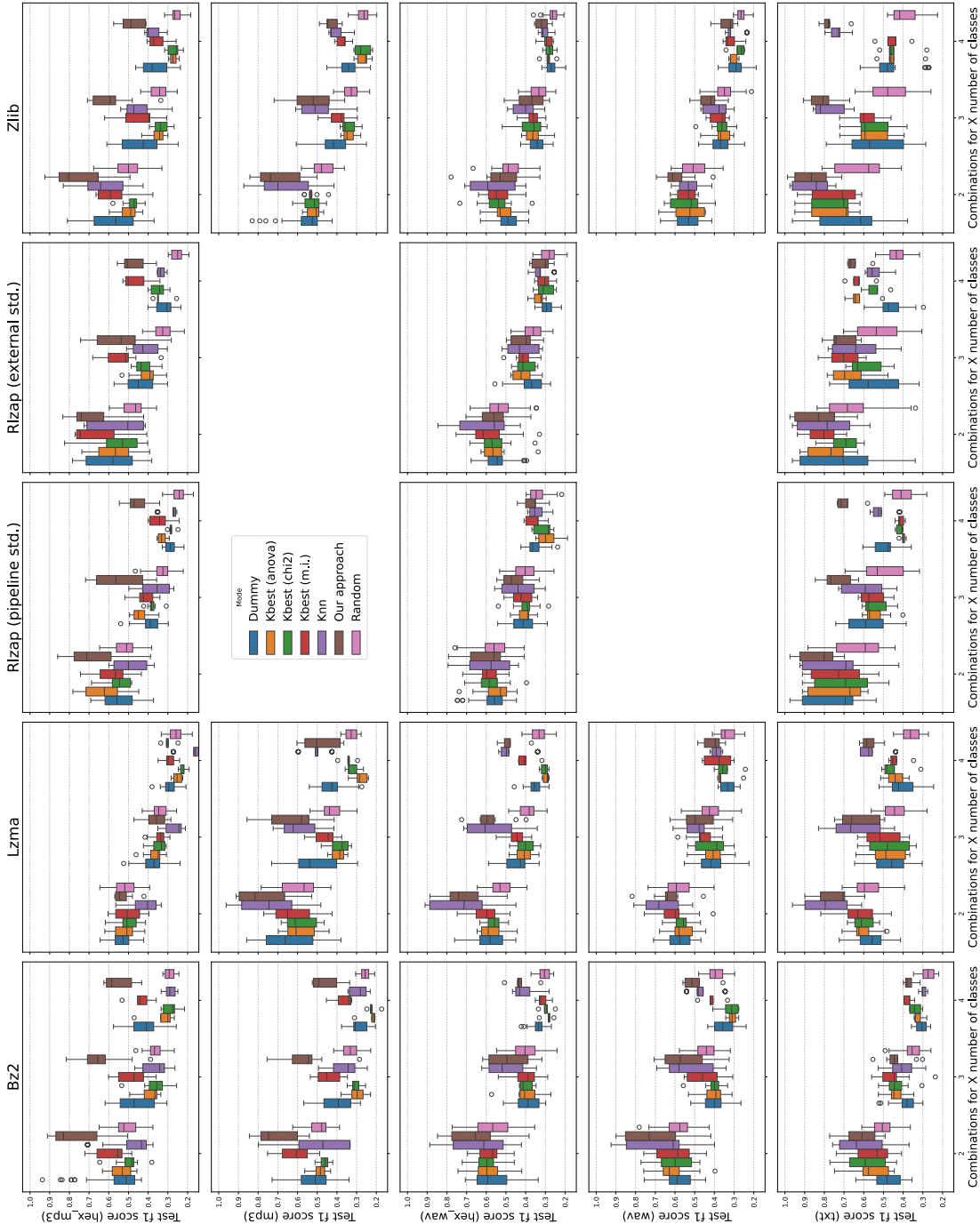}
  \caption{\textbf{Very hard dataset result distributions} for each class
    combination using our approach and alternative methods. Each row represents
    a different file format (txt, wav, mp3, hex\_wav, and hex\_mp3), while each
    column corresponds to one of the five compressors evaluated (zlib, bz2,
    lzma, rlzap with external standardization, and rlzap with pipeline
    standardization). Due to incompatibilities already discussed in
    Section~\ref{subsec:rlzap}, the rlzap methods do not include results for the
    Wav and Mp3 formats. All plots show test F1 score distributions across all
    class combinations of size two, three, and four. ``Kbest'' modes refer to
    feature selection alternatives to our approach, while ``Dummy'' and
    ``Random'' modes represent the baseline methods previously described.
    ``Knn'' denotes a Nearest Neighbors classifier computing the Euclidean
    distance over the original dissimilarity matrix. Each parameter under our method
    were selected through a grid search, and the number of features in
    alternative and random modes was chosen via simple iteration. Additionally,
    as previously introduced, the knn cases uses all the samples of the training
    set while the other methods use a subset of those. }
  \label{fig:boxplot_very_hard_num_classes}
\end{sidewaysfigure}

Figure~\ref{fig:boxplot_very_hard_num_classes} illustrates the performance
across different compression methods (zlib, bz2, lzma, and two variants of rlzap), evaluated on multiple file encodings: plain text (txt), binary waveform
(wav), mp3, and their hexadecimal representations (hex\_wav and hex\_mp3). As
discussed in Section~\ref{subsec:rlzap}, rlzap is incompatible with the wav and
mp3 formats directly, and thus operates only on their corresponding hexadecimal
representations. This figure also distinguishes between standardization modes
applied to rlzap (as described in previous experiments): using an external set
of samples or one from the training set.

The results, summarized in Table~\ref{tab:very_hard_results}, indicate a marked
degradation in performance compared to previous experiments (see
Table~\ref{tab:medium_gridresults} and
Figure~\ref{fig:boxplot_medium_num_classes}). While the best-performing method
in the earlier textual settings exceeded 0.9 F1 score for some class
combinations, the highest performance observed in this scenario—achieved by zlib
on txt files—barely approaches 0.77. In contrast, rlzap, under all
configurations, fails to surpass 0.70 across all five classes.

These results are further contextualized by the figure showing performance
across all class combinations of size two, three, and four (see
Figure~\ref{fig:boxplot_very_hard_num_classes}). The boxplots clearly show a
downward trend in classification performance as the number of target classes
increases. Notably, some methods exhibit more abrupt degradation than others.
Despite the overall decline in accuracy, our method maintains a performance
level that is generally higher than the baseline provided by knn.
While knn occasionally outperforms our approach in simpler
configurations—specifically those involving only two or three classes—the
performance distribution in these scenarios tends to be slightly skewed in favor
of knn. However, as class complexity increases, the margin between the two
approaches widens, and our method consistently outperforms knn.

Focusing specifically on the four-class combinations in
Figure~\ref{fig:boxplot_very_hard_num_classes}, our approach exhibits a
particularly clear advantage. Across the 21 configurations considered, there is
only a single case in which knn attains a higher maximum F1 score than our
method, and only three cases in which the overall F1 distribution of knn can be
regarded as better than that of our approach. For the best-performing
configuration in this scenario (for instance, compression with zlib on
plain-text representations), our approach outperforms knn both in terms of the
maximum F1 achieved and in the overlap between the corresponding F1-score
distributions. For other class-combination sizes (two and three classes), our
approach remains competitive or superior. Feature selection via
kbest does not provide substantial improvement in this scenario. In many cases,
the selected features perform comparably to, or only slightly better than,
random subsets, with some minor exceptions of kbest under mutual information that provide slightly better results than the other two kbest methods.

Finally, we note that the evaluation framework for this experiment is entirely
dependent on the ability of each method to identify the most representative
fragment per test audio (as the decission is made by the ``most representative
sample of each audio file''). This procedure is uniformly applied across all
methods and formats, and thus does not introduce bias in favor of any particular
approach. Nevertheless, it is conceivable that a more refined selection strategy
could improve overall accuracy, although such optimizations have been
deliberately omitted in favor of maintaining methodological simplicity and
ensuring fair comparison between approaches.

\section{Discussion}
\label{sec:discussion}

The three experimental scenarios presented in this work offer increasing levels
of complexity, allowing us to investigate the robustness and behavior of
compression-based analysis under diverse conditions. Each case provides distinct
insights into the dynamics between data characteristics, compression methods,
and the structure of the resulting similarity space.

In the simplest setting, involving short stories from two authors, we observed
high classification performance across nearly all methods, including random and
dummy baselines. This is consistent with the relatively trivial nature of the
task and the strong coherence within classes. One interesting point is the
number of objects used in the process. While our method managed to produce good
results using a limited number of samples, the alternatives benefited highly
from being able to use more objects. This could imply that, as our method
focuses on finding "the optimal" samples for the current context, it doesn't
benefit that much from having access to more samples, as likely the best ones are
already selected. This characteristic of our method has not been explored in
this work, but could be a good line of research for future work. Nonetheless,
our method was able to differentiate itself from these baselines by maintaining
consistency across compressors and feature subsets, and even improving performance
in some cases. While rlzap without standardization underperformed significantly
across all methods, its standardized variants achieved remarkably strong results.
This suggests that even in low-complexity settings, the choice of normalization
strategy plays a decisive role when dealing with NRC.

The second experiment, involving a six-class text dataset and its combinatorial
subsets, introduced controlled complexity increases in small increments. This
design proved crucial: by incrementally increasing the number of classes, we
were able to detect the fragility of certain baselines. Both dummy and random
methods exhibited steep performance drops, confirming that complexity—not noise
alone—challenges compressive similarity. Our method, alongside knn and the
better-performing kbest variants, preserved classification accuracy more
effectively. Among these, external standardization in rlzap emerged as a
particularly robust strategy, offering greater generality and improved
clustering quality compared to pipeline-based standardization. Notably, knn
suffered a surprising performance drop in combination with zlib (as it did in
the first experiment), likely due to subtle misalignments between zlib's
compression patterns and the structure expected by knn when using Euclidean
metrics. While no definitive conclusion was drawn, the result underscores the
sensitivity of classical classifiers (knn) to the geometry induced by
compressive transformations.

The third and most complex experiment utilized a real-world audio dataset with
audios from five different species, fragmented to allow comparison across
heterogeneous recordings. This case exposed the limitations of compression
methods when faced with unstructured and noisy data. Performance across all
methods dropped substantially compared to previous scenarios. Still, our method
consistently produced competitive F1 scores, particularly when paired with bz2,
zlib, or rlzap applied to appropriate file formats. In some specific
configurations, such as bz2 + hex\_mp3 or zlib + txt, our method showed clear
improvements over all alternatives with minimal overlap in performance
distributions.

Interestingly, the silhouette coefficient proved unreliable in this final case,
producing uniformly negative scores. This can be attributed to a combination of
factors: the use of only the most representative test fragment per audio, the
sparsity of true neighborhoods in high-complexity data, and the overall noise
and label variability present in the dataset. Although silhouette remains a
useful metric in controlled environments, its application here failed to yield
informative clustering structure and was thus deprioritized in the
interpretation of results. In the Very Hard setting, clustering-based scores are
markedly weaker than in the Easy and Medium datasets, indicating that global
cluster structure degrades faster than the supervised signal exploited by the
classifier in this scenario.

Altogether, these experiments reinforce the potential of our methodology not
only to provide a consistent analytical framework, but also to remain robust
under varied and challenging conditions. They also emphasize the importance of
moving away from ad hoc solutions, advocating instead for more flexible,
modular, and interpretable approaches to compression-based learning.
\section{Conclusions}
\label{sec:conclusions}

This work has introduced a supervised context-steering methodology for deriving
task-oriented embeddings from compression dissimilarities, providing a flexible and
systematic framework for building features and enabling downstream tasks. In
contrast to traditional uses of the NCD, which
rely mostly on the emergent structure of pairwise similarities or ad hoc
approaches, our approach allows us to shape the problem towards a given goal
taking advantage from the context extracted from data. We refer to this property
as \textit{context steering}—the ability to influence the resulting structure by
means of sample and feature selection on an autonomous manner, rather than
depending solely on the similarities found by compression dissimilarities.

Beyond this conceptual contribution, we proposed several methodological
innovations that enhance the efficiency and versatility of the framework. We
introduced the use of hierarchical clustering with Euclidean distances as a
scalable alternative to traditional clustering techniques applied directly to
NCD matrices. We also explored the application of rlzap with NRC as a
competitive compression-based similarity, particularly suitable for our
distance-as-feature formulation. Furthermore, our validation procedure was
designed to be domain-agnostic, enabling robust comparisons across heterogeneous
data types, compression schemes, and classification scenarios.

These contributions were assessed through a series of experiments of increasing
complexity. In the simplest setting, involving binary classification of textual
documents, even naive methods performed reasonably well. Nevertheless, our
method consistently delivered superior performance across compressors while
using lighter embeddings (\textit{i.e.} using less objects to transform data into the new
feature space). As additional classes were introduced, our approach maintained
this advantage, outperforming other methods such as knn and
kbest-based alternatives—especially when standardized relative
compression dissimilarities were employed.

Finally, unlike nearest-neighbor models such as knn, which require retaining the
entire dataset at inference time, our method relies on a reduced and structured
representation. This inherently yields greater efficiency in terms of storage
and computational cost. Moreover, since the second step of our sample selection
(Section~\ref{subsubsec:second_step}, step 2 in
Figure~\ref{fig:method_overview}) process yields cluster-level scores, our
method implicitly defines a prioritization mechanism. This enables an
interesting trade-off mechanism to be able to choose between accuracy and
inference speed, by limiting the number of clusters or elements considered.
Further analysis of this accuracy-speed compromise constitutes an interesting
direction for future exploration.

Thus, the main contributions of our work can be summarized as follows:

\begin{itemize}
  \item We propose Context Steering, a methodology for working with compression
  distances in supervised settings that defines a task-oriented feature space by
  using selected objects as references.

  \item We conduct an extensive experimental study on three datasets of
  increasing difficulty—from binary text authorship to heterogeneous multi-class
  audio—showing that Context Steering yields competitive embeddings from both
  NCD and NRC, and that these embeddings remain useful for downstream analysis
  across datasets of increasing difficulty, with clustering-based evidence being
  strongest in the first two settings and less reliable in the most complex
  audio scenario.

  \item We show that the relational context induced by clustering compression
  distances can be actively reshaped through sample and feature selection. In
  particular, we turn parts of the dissimilarity matrix into a task-oriented
  embedding.

  \item We compare classical compressors and the NCD itself with rlzap applied
    through NRC, as a suitable alternative given the feature-based
    interpretation of distances in our method.

  \item We present a validation framework applicable to heterogeneous data and
  multiple compression algorithms without requiring domain-specific adjustments,
  allowing the model to be evaluated across various contexts, problem settings,
  and parameter choices.

  \item We demonstrate the robustness and scalability of our approach through a
  suite of increasingly complex experiments, including a real-world scenario
  involving multi-class audio data with mixed file formats.

  \item We show that the method yields more consistent and robust results than
  alternatives such as knn and kbest-based feature selection,
  and that it outperforms them in several cases by notable margins.

  \item We demonstrate that the model does not require access to the full
  dataset at inference time, enabling its use in inductive classification
  settings, unlike transductive methods such as the quartet-based
  method usually paired with compression dissimilarities.

  \item We introduce a cluster-level scoring mechanism that enables the prioritization of representative samples, paving the way for future work aimed at mitigating the well-known efficiency issues of compression-based dissimilarities.
\end{itemize}

Altogether, these contributions illustrate how compression-based similarity can
be shifted from passive discovery to active control, enabling not only richer
analysis but also more efficient and interpretable compression-based methods.
\section*{Acknowledgments}
This work was supported by Grant PID2023-149669NBI00 (MCIN/AEI and ERDF – ``A way of making Europe'')
\section*{Declaration of generative AI and AI-assisted technologies in the manuscript preparation process}

During the preparation of this work, the author(s) used ChatGPT (OpenAI) to improve the language, clarity, and readability of the manuscript. After using this tool, the author(s) reviewed and edited the manuscript as needed and take(s) full responsibility for the content of the published article.

\bibliographystyle{plain}
\bibliography{references}

\section{Notation}
\label{sec:notation}

We summarize here the notation used throughout the paper for quick reference.

\paragraph{Summary of symbols.}
\begin{itemize}
  \item $X=\{x_i\}_{i=1}^{N}$: set of data objects.
  \item $Y=\{y_i\}_{i=1}^{N}$, $y_i\in\{1,\dots,K\}$: class labels; $K$ is the total number of classes.
  \item $\mathcal{C}_k=\{i\mid y_i=k\}$: index set of objects belonging to class $k$.
  \item $C(\cdot)$: compression operator, compressed representation, or compressed size. It can represent $C(x)$ and $C(xy)$, where $x$ is a string and $xy$ denotes the concatenation of $x$ and $y$; it can also denote conditional compression, such as $C(x\mid y)$, or exclusive conditional/relative compression, such as $C(x\Vert y)$, where $y$ is the reference used to compress $x$.
  \item $d(\cdot,\cdot)$: distance or dissimilarity operator, often instantiated as a compression-dissimilarity operator such as NCD or NRC.
  \item $\mathcal{M}\in\mathbb{R}^{N\times N}$: full pairwise dissimilarity matrix, often a compression-dissimilarity matrix, with entries $m_{ij}=d(x_i,x_j)$.
  \item $\delta_{\mathrm{eucl}}(\cdot,\cdot)$: Euclidean distance between behavior profiles or, more generally, between two vectors; for example, $\delta_{\mathrm{eucl}}(u,v)=\sqrt{\sum_{i=1}^{N}(u_i-v_i)^2}$. It may also be applied between a vector and a matrix as a vectorized distance computation.
  \item $\delta_{\mathrm{alt\_eucl}}(\cdot,\cdot)$: optional modified or blockwise Euclidean alternative (Section~\ref{subsubsec:add_step}).
  \item $E\in\mathbb{R}^{N\times N}$: Euclidean matrix computed from $\mathcal{M}$, with entries $E_{p,q}=\delta_{\mathrm{eucl}}(\mathcal{M}[p,:],\mathcal{M}[q,:])$, or using $\delta_{\mathrm{alt\_eucl}}$ when the alternative is enabled.
  \item $E_k = E[\mathcal{C}_k,\mathcal{C}_k]$: within-class restriction of $E$, equivalently the within-class submatrix associated with class $k$.
  \item $\mathrm{AHC}_{\mathrm{Linkage}}(\cdot)$: hierarchical clustering linkage operator; Ward linkage is used by default.
  \item $\mathcal{T}_k := \mathrm{AHC}_{\mathrm{Linkage}}(E_k)$: class-wise dendrogram built from $E_k$ using the selected hierarchical clustering linkage criterion.
  \item $\mathcal{P}(\mathcal{T})$: partition-generation operator over a dendrogram $\mathcal{T}$; it denotes the set of candidate partitions obtainable from $\mathcal{T}$.
  \item $\mathrm{SC}(\cdot)$: Silhouette Coefficient functional used for partition or cluster quality evaluation.
  \item $\mathcal{G}_k^\ast$: best partition of $\mathcal{T}_k$ under the Silhouette Coefficient criterion.
  \item $h(\mathcal{G},\tau)$: cluster-selection operator applied after a partition $\mathcal{G}$ is available.
  \item $\tau$: number of retained clusters in the top-$\tau$ selection strategy, with $\tau\in\mathbb{N}$; alternatively, a non-integer value of $\tau$ denotes an above-average selection mode.
  \item $G=\{G_j\}_{j=1}^{J}$: global set or subset of selected coherence groups, typically obtained by collecting $h(\mathcal{G}_k^\ast,\tau)$ across classes and reindexing the selected clusters.
  \item $\mathcal{L}_j = \mathcal{M}[G_j,G_j]$: within-cluster submatrix or list of compression dissimilarities restricted to members of $G_j$; it is used for reference selection and weighting.
  \item $\Delta_j(x,\Theta)$: average-distance measurement function for cluster $j$, measuring the average Euclidean distance from $x$ to the elements in $\Theta\subseteq G_j$ using $\mathcal{L}_j$ (Eq.~\ref{eq:step3_centroid_score}).
  \item $F$: number of reference objects selected per cluster or coherence group (hyperparameter).
  \item $\mathrm{ref\_mode}\in\{\mathrm{close},\mathrm{far}\}$: inner-reference selection policy, distinguishing between a central reference-selection strategy and a covering or farthest-reference strategy.
  \item $R_j=\{r_j^i\}_{i=1}^{F}\subseteq G_j$: set of reference objects selected for group $G_j$.
  \item $\omega_j^i\in\mathbb{R}^{|G_j|}$: weight vector associated with reference $r_j^i$, encoding its similarity to the other elements within cluster $G_j$ (Eq.~\ref{eq:step3_weight}).
  \item $V_{j,i}(x)\in\mathbb{R}^{|G_j|}$: reference-derived embedding coordinate or reference-conditioned dissimilarity vector of $x$, associated with group $G_j$ through reference $r_j^i$ (Eq.~\ref{eq:values}).
  \item $\phi_f(x)\in\mathbb{R}^{FJ}$: final task-oriented embedding map obtained by aggregating each $V_{j,i}(x)$ and concatenating the resulting coordinates across groups and references.
\end{itemize}

\paragraph{Indexing and slicing conventions.}
Matrix indexing follows the standard row-major $[\text{row},\text{column}]$
convention. The colon symbol $:$ denotes full dimension traversal. For example,
$\mathcal{M}[i,:]$ selects the entire $i$-th row and $\mathcal{M}[:,j]$ selects the entire $j$-th
column. For a set of indices $S$, $\mathcal{M}[i,S]$ denotes restriction of row $i$ to
columns in $S$, and $\mathcal{M}[S,S]$ denotes the square submatrix restricted to rows and
columns in $S$.

\appendix

\section{Reference selection in Context Steering}
\label{app:ref_selection}

This appendix reports the inner-reference selection procedure used in
Step~3.a (Section~\ref{subsubsec:third_step}), following the formulation
introduced in Chapter~12. Let $G_j$ be one of the selected coherence groups and
let $\mathcal{L}_j = \mathcal{M}[G_j,G_j]$ denote the within-group restriction of the
compression-dissimilarity matrix.

\paragraph{Centrality functional.}
For any non-empty $\Theta\subseteq G_j$ and any $x\in\Theta$, we define the
average (Euclidean) distance from $x$ to the remaining members of $\Theta$ in the
row space of $\mathcal{L}_j$ as
\begin{equation}
  \Delta_j(x,\Theta)
    = \frac{1}{|\Theta \setminus \{x\}|}
      \sum_{x' \in \Theta \setminus \{x\}}
      \delta_{\text{eucl}}\!\big(\mathcal{L}_j[x,:],\mathcal{L}_j[x',:]\big).
  \label{eq:app_Delta}
\end{equation}
Small values of $\Delta_j(x,\Theta)$ indicate that $x$ is central
(centroid-like) in $\Theta$, whereas large values indicate that $x$ is
peripheral.

\paragraph{Recursive selection sets.}
We define a family of recursive selection sets $S^n_{j,t}$, parameterized by
$n\in\{0,1\}$ and the iteration index $t\in\mathbb{N}$, by first setting
\begin{equation}
  S^n_{j,1}
  = \bigg\{ \arg\min_{x \in G_j} (-1)^n \,\Delta_j(x,G_j) \bigg\}.
  \label{eq:app_S_base}
\end{equation}
For $n=0$, this yields the most central element of $G_j$; for $n=1$, it yields
the most peripheral element (largest $\Delta_j$).

For $t\ge 2$, we extend the set recursively as
\begin{equation}
  S^n_{j,t}
  =
  S^n_{j,\,n(t-2)+1}
  \;\cup\;
  \arg\min_{\substack{\mathcal{S}\subseteq G_j\setminus S^{n}_{j,n(t-2)+1} \\
                      |\mathcal{S}|=(t-2)(1-n)+1}}
  \sum_{x \in \mathcal{S}} (-1)^n \,\Delta_j\!\big(x, S^n_{j,t-1}\big).
  \label{eq:app_S_rec}
\end{equation}

\paragraph{Close vs far modes.}
Using the recursive sets above, the reference pool after $t$ iterations is
\begin{equation}
  R^{\mathrm{mode}}_{j,t}
  =
  \begin{cases}
    S^0_{j,t},   & \mathrm{mode}=\mathrm{close},\\
    S^{1}_{j,t},& \mathrm{mode}=\mathrm{far}.
  \end{cases}
  \label{eq:app_R_mode_t}
\end{equation}
The \texttt{close} mode selects the most central (medoid-like) elements of $G_j$,
whereas the \texttt{far} mode selects peripheral/covering elements intended to be
maximally spread.

Given the hyperparameter $F\ge 1$ (number of references per group) and the
configuration $\mathrm{ref\_mode}\in\{\mathrm{close},\mathrm{far}\}$, the
references for group $G_j$ are defined as
\begin{equation}
  R_j = R^{\mathrm{ref\_mode}}_{j,F}\subseteq G_j,
\end{equation}
and the global reference set is $R=\bigcup_{j=1}^{J} R_j$.

\begin{algorithm}[H]
  \footnotesize
\begin{algorithmic}[1]
\Require Compression-dissimilarity matrix $\mathcal{M} \in \mathbb{R}^{N \times N}$, class labels $y \in \{1,\dots,K\}^N$, hyperparameters $\Theta=(\mathrm{Linkage},\tau,F,\mathrm{ref\_mode},f)$
\Ensure Embedding map $\phi_f : \{1,\dots,N\} \rightarrow \mathbb{R}^d$ and embedded matrix $Z \in \mathbb{R}^{N \times d}$

\If{\texttt{use\_modified\_eucl}} \Comment{Step 1}
    \State $\delta(p, q) \gets \delta_{\mathrm{alt\_eucl}}(p, q)$ \Comment{cf. (\ref{eq:our_euclidean})}
\Else
    \State $\delta(p, q) \gets \delta_{\mathrm{eucl}}(p, q)$ \Comment{cf. (\ref{eq:euclidean})}
\EndIf

\State $E \gets \texttt{zeros}(N,N)$
\For{$p = 1$ to $N$}
    \For{$q = 1$ to $N$}
        \State $E_{p,q} \gets \delta\!\left(\mathcal{M}[p,:],\mathcal{M}[q,:]\right)$
    \EndFor
\EndFor
\Comment{$E$: behavior-similarity matrix}

\State $G \gets \emptyset$
\For{$k = 1$ to $K$} \Comment{Step 2}
    \State $\mathcal{C}_k \gets \{ i \mid y_i = k \}$
    \State $E_k \gets E[\mathcal{C}_k,\mathcal{C}_k]$
    \State $\mathcal{T}_k \gets \mathrm{AHC}_{\mathrm{Linkage}}(E_k)$
    \State $\mathcal{G}_k^\ast \gets \mathrm{SelectBestPartition}(\mathcal{T}_k)$ \Comment{cf. (\ref{eq:group_of_clusters})}
    \State $G \gets G \cup h(\mathcal{G}_k^\ast,\tau)$ \Comment{cf. (\ref{eq:cluster_selection})}
\EndFor \Comment{$G$: set of selected clusters across all classes}

\State $R \gets \emptyset$
\For{$j = 1$ to $J$} \Comment{Step 3.a}
    \State $\mathcal{L}_j \gets \big[m_{x,y}\big]_{x,y \in G_j}$
    \For{$t = 1$ to $F$}
        \If{$\mathrm{ref\_mode}=\mathrm{close}$} \Comment{cf. (\ref{eq:app_R_mode_t})}
            \State $R_j^t \gets \mathcal{S}_{j,t}^{0}$ \Comment{cf. (\ref{eq:app_S_rec})}
        \Else
            \State $R_j^t \gets \mathcal{S}_{j,t}^{1}$ \Comment{cf. (\ref{eq:app_S_rec})}
        \EndIf
        \State $R_j \gets R_j \cup \{r_j^t\}$
    \EndFor
\EndFor
\Comment{$R$: set of references associated with the selected clusters in $G$}

\For{$j = 1$ to $|G|$} \Comment{Step3.b and 3.c}
    \For{$i = 1$ to $F$}
      \State $\omega^i_{j} \gets 1-\mathrm{norm}_{[0,1]} \Big(\delta\big(\mathcal{L}_{j}[r^i_{j},:],\mathcal{L}_{j}\big) \Big) $ \Comment (cf. \ref{eq:step3_weight})
      \State $V^i_{j}(x) \gets \delta \!\Big(\mathcal{M}[x,G_j],\; \mathcal{L}_j \cdot \mathrm{diag}(\omega^i_j) \Big) $ \Comment cf. (\ref{eq:values})
        \State $\phi^{\,i}_{j}(x) \gets f(V_j^i(x))$ \Comment cf. (\ref{eq:phi})
    \EndFor
\EndFor

\State \Return $\phi$
\end{algorithmic}

\caption[Context Steering algorithm]{\textbf{Context Steering algorithm.} Pseudocode for transforming a compression-dissimilarity matrix into a task-oriented embedding usable for classification and clustering.}
\label{alg:context_steering}
\end{algorithm}

\begin{algorithm}

\begin{algorithmic}[1]
\Require Feature matrix $X$, distance metric $\texttt{metric}$, linkage criterion $\texttt{linkage}$
\Ensure Dendrogram of merges
\If{$\texttt{metric} \ne \texttt{'precomputed'}$}
    \State $\Sigma \gets \texttt{get\_distance\_matrix}(X, \texttt{metric})$
\Else
    \State $\Sigma \gets$ precomputed matrix
\EndIf
\State $C \gets \{ \{x_1\}, \{x_2\}, \dots, \{x_n\} \}$
\State $\texttt{dendrogram} \gets [\ ]$
\While{$|C| > 1$}
    \State $(s, t) \gets \texttt{get\_closest\_pair}(C, \Sigma)$
    \State $u \gets \texttt{join\_clusters}(s, t)$
    \State $\texttt{dendrogram.append}((s, t, \Sigma[s,t]))$
    \State $C \gets C \setminus \{s, t\}$
    \ForAll{$v \in C$}
        \State $\Sigma[v,u] \gets \texttt{get\_cluster\_distance}(v, u, \Sigma, X, \texttt{linkage})$
    \EndFor
    \State $C \gets C \cup \{u\}$
    \State $\Sigma \gets \texttt{remove\_rows}(\Sigma, \{s, t\})$
\EndWhile
\State \Return $\texttt{dendrogram}$
\end{algorithmic}
\caption{\textbf{Agglomerative Hierarchical Clustering Algorithm}: The algorithm iteratively merges the two closest clusters according to a linkage criterion, updating the distance matrix at each step. The process continues until a single cluster remains or a stopping condition is met (e.g., number of clusters or distance threshold). Depending on the linkage method (e.g., single, complete, average), the way inter-cluster distances are computed varies. The resulting structure is a dendrogram representing the nested hierarchy of clusters.}
\label{alg:hca}
\end{algorithm}
\end{document}